\documentclass[11pt]{article}

\PassOptionsToPackage{table}{xcolor}
\usepackage[preprint]{acl}
\usepackage{times}
\usepackage{latexsym}
\usepackage[T1]{fontenc}
\usepackage[utf8]{inputenc}
\usepackage{microtype}
\usepackage{inconsolata}
\usepackage{amsmath,amssymb,amsfonts}
\usepackage{graphicx}
\usepackage{booktabs}
\usepackage{subcaption}
\usepackage{placeins} %
\usepackage{tabularx}
\usepackage[most]{tcolorbox}
\usepackage{pifont}
\usepackage{fontawesome5}
\definecolor{promptbg}{HTML}{E8F1E4}
\definecolor{truthbg}{HTML}{F2F2F2}
\definecolor{wrongbg}{HTML}{FBE9E7}
\definecolor{rightbg}{HTML}{E4EEF8}
\definecolor{panelframe}{HTML}{9E9E9E}
\definecolor{titlebg}{HTML}{F0F0F0}
\definecolor{errred}{HTML}{C0392B}
\definecolor{okgreen}{HTML}{1E7B34}

\usepackage{tikz}

\definecolor{eaGreen}{HTML}{008300}
\definecolor{eaAqua}{HTML}{1BAF7A}
\definecolor{eaYellow}{HTML}{EDA100}
\definecolor{eaMagenta}{HTML}{E87BAA}
\definecolor{eaViolet}{HTML}{4A3AA7}
\definecolor{eaRed}{HTML}{E34948}
\definecolor{eaBlue}{HTML}{2A78D6}

\definecolor{cGPT55}{HTML}{6A4C93}
\definecolor{cKimi}{HTML}{B5179E}
\definecolor{cClaude}{HTML}{E07A5F}
\definecolor{cGemini}{HTML}{0091A7}
\definecolor{cGemma}{HTML}{7A9E29}
\definecolor{cGPT56}{HTML}{8C6239}
\definecolor{cCodex}{HTML}{C9971B}
\definecolor{cChandra}{HTML}{1B7F3B}
\definecolor{cQwen25}{HTML}{1F5FA9}
\definecolor{cQwen35}{HTML}{C1272D}

\definecolor{gridline}{HTML}{C8C8C8}
\definecolor{gridedge}{HTML}{2B2B2B}
\definecolor{axistext}{HTML}{1A1A1A}

\colorlet{headerbg}{titlebg}
\colorlet{failurebg}{wrongbg}
\colorlet{successbg}{rightbg}

\colorlet{cQwen150}{cQwen25}
\colorlet{cQwen50}{cQwen25!50!white}
\colorlet{cChandra150}{cChandra}
\colorlet{cChandra50}{cChandra!50!white}
\colorlet{cAngelina}{black!55}

\newtcolorbox{modelpanel}[1]{%
  enhanced,
  arc=2pt,
  colback=white,
  colframe=panelframe,
  boxrule=0.7pt,
  left=4pt,
  right=4pt,
  top=3pt,
  bottom=3pt,
  title=#1,
  fonttitle=\bfseries\footnotesize,
  colbacktitle=titlebg,
  coltitle=black,
  toptitle=2pt,
  bottomtitle=2pt
}

\newtcolorbox{tintbox}[1]{%
  colback=#1,
  colframe=#1,
  boxrule=0pt,
  arc=1.5pt,
  left=5pt,
  right=5pt,
  top=4pt,
  bottom=4pt,
  before skip=2pt,
  after skip=2pt
}

\newcommand{\xmark}{\textcolor{errred}{\ding{55}}}
\newcommand{\cmark}{\textcolor{okgreen}{\ding{51}}}

\definecolor{sampleframe}{HTML}{D2D7DC}
\definecolor{samplebg}{HTML}{FAFBFC}
\definecolor{samplelabel}{HTML}{F0F3F5}
\definecolor{sampletext}{HTML}{4A4A4A}

\newtcolorbox{samplecard}[1][]{%
  enhanced,
  colback=samplebg,
  colframe=sampleframe,
  boxrule=0.55pt,
  arc=3pt,
  outer arc=3pt,
  left=5pt,
  right=5pt,
  top=5pt,
  bottom=5pt,
  boxsep=0pt,
  before skip=0pt,
  after skip=0pt,
  #1
}

\newtcolorbox{apppanel}[2][]{%
  enhanced,
  colback=white,
  colframe=panelframe,
  colbacktitle=titlebg,
  coltitle=black,
  fonttitle=\bfseries,
  title={\strut #2},
  boxrule=0.6pt,
  arc=2pt,
  left=6pt,
  right=6pt,
  top=6pt,
  bottom=6pt,
  toptitle=3pt,
  bottomtitle=3pt,
  #1
}

\newtcolorbox{appfillbox}[2][]{%
  colback=#2,
  colframe=#2,
  boxrule=0pt,
  arc=0pt,
  left=8pt,
  right=8pt,
  top=6pt,
  bottom=6pt,
  #1
}

\newif\ifshownotes
\shownotesfalse

\title{Learning to Read Braille: Benchmarking and Improving Braille Recognition Capabilities of VLMs}
\author{%
  \textbf{Kartik Jangra}$^{1,2\dagger}$\quad
  \textbf{Atharva Gawane}$^{1,2\dagger}$\quad
  \textbf{Sidharth Pulipaka}$^{1}$\quad
  \textbf{Hemachandran Karah}$^{2}$\quad
  \textbf{Raj Dabre}$^{1,2}$\\[6pt]
  $^{1}$Nilekani Centre at AI4Bharat\quad
  $^{2}$Indian Institute of Technology, Madras\\[6pt]
  {\small
  \faGithub\ \href{https://github.com/kartik0806/BrailleBench}{\textcolor{blue}{github.com/kartik0806/BrailleBench}}
  \quad
  \raisebox{-0.2em}{\includegraphics[height=1em]{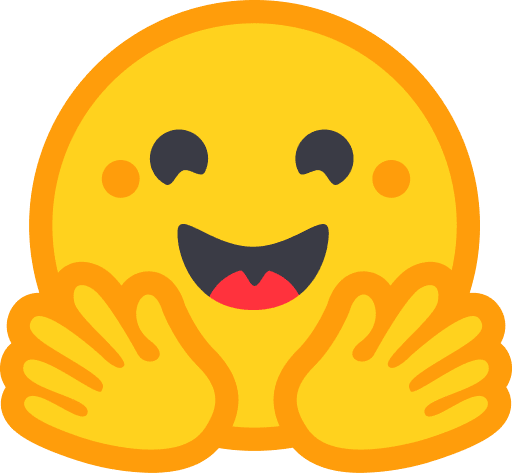}}\ \href{https://huggingface.co/ai4bharat/BrailleQwen-2B}{\textcolor{blue}{huggingface.co/ai4bharat/BrailleQwen-2B}}}
}

\begin{document}
\maketitle
{\renewcommand{\thefootnote}{$\dagger$}\footnotetext{Equal contribution.}}

\begin{abstract}
Despite rapid progress in visual understanding and document transcription, modern vision-language models (VLMs) struggle with a fundamental accessibility task: reading Braille from images. We introduce BrailleBench, a manually curated benchmark of 259 real-world Grade-1 and Grade-2 Braille images. We evaluate proprietary, open-weight, OCR-specialized, and tool-assisted multimodal systems and find that standalone general-purpose models perform poorly at directly transcribing Braille images, whereas tool-assisted systems perform much better. We design a synthetic Braille generation pipeline with diverse layouts and image perturbations. A 2B VLM model trained on our synthetic data reaches 87\% exact match on Grade-1 and 61\% on Grade-2 BrailleBench, outperforming standalone VLMs and achieving the highest Grade-2 exact match score among the evaluated systems. We further show that high accuracy on synthetic validation data does not guarantee high accuracy on real photographs in our experiments: models trained on prior synthetic data reach up to 96\% on their synthetic validation set but at most 3.26\% on Grade-1 BrailleBench. Our results identify an accessibility-relevant capability gap in multimodal systems and show that targeted synthetic supervision can improve real-world Braille transcription.

\end{abstract}

\section{Introduction}
\label{sec:introduction}

\begin{figure*}[t]
    \centering
    \includegraphics[width=\textwidth]{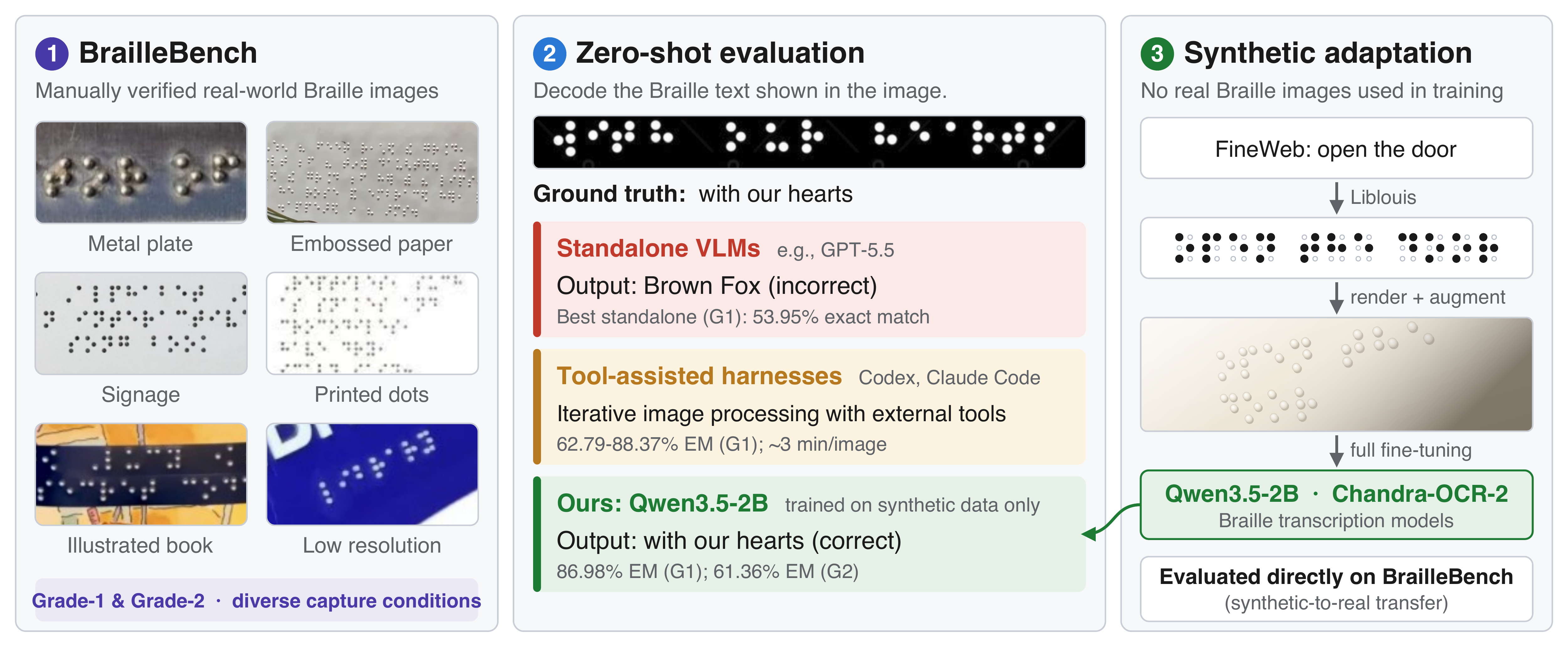}
    \caption{Overview of our work. (1)~BrailleBench contains manually verified
    real-world Braille images captured on diverse materials and under diverse
    conditions. (2)~Standalone VLMs largely fail to transcribe them, often
    producing fluent but unrelated text, while tool-assisted harnesses succeed
    at a high inference cost. (3)~Models fine-tuned only on synthetically
    generated Braille images transcribe real-world Braille with a single model call,
    without a tool loop. EM denotes exact match; G1 and G2 denote Grade-1 and Grade-2.}
    \label{fig:overview}
\end{figure*}

Modern vision-language models (VLMs) can read documents, handwriting,
screenshots, charts, and scene text, and recent systems combine them with
agentic tool use for multi-stage visual processing
\cite{openai2026gpt5.6,anthropic2026opus48,moonshotai2026kimik26}.
Strong results on benchmarks such as ChartQA \cite{masry2022chartqa} and
OmniDocBench \cite{ouyang2025omnidocbench}, however, do not imply uniform
capability across writing systems. Braille provides an accessibility-relevant test of whether these capabilities
extend beyond conventional printed text.

Globally, at least 2.2 billion people live with vision impairment
\cite{who2019worldvision}, including 43 million people who are blind and
295 million with moderate to severe vision loss in the 2020 estimates
\cite{iapb2021visionatlas}. For many of them, Braille remains essential to
literacy, education, and access to information \cite{unworldbrailleday}.
Braille is also regularly encountered by sighted people who do not read it
fluently, such as teachers checking students' work or family members
reading a note. A model that reliably reads Braille from an image could
therefore serve as a bridge between Braille and non-Braille readers in
education, at home, and in public life. 

Reading Braille from images differs from conventional OCR since a
character is defined entirely by the presence and relative position of
raised dots within a small cell, often embossed on low-contrast surfaces
whose appearance depends on illumination, material, perspective, and
resolution. Recognition therefore relies on fine-grained counting and
geometric reasoning, on which current VLMs are known to struggle
\cite{guo2025your}. Furthermore, Braille is written in two forms:
Grade-1 spells text letter by letter, while Grade-2 adds contractions
whose reading depends on context (Section~\ref{subsec:data_collection}).

We introduce \textbf{BrailleBench}, a manually verified benchmark of 259
real-world Braille images (215 Grade-1 and 44 Grade-2) spanning diverse
text lengths, materials, and capture conditions, and use it to evaluate
proprietary, open-weight, OCR-specialized, and tool-assisted systems
(Figure~\ref{fig:overview}). Standalone VLMs perform poorly: on Grade-1,
the strongest, GPT-6.1 Sol, reaches only 53.95\% exact match, and the
others at most 11.16\%. Tool-assisted harnesses do far better, with
Claude Code (Opus 5.5 High) reaching 88.37\%, but they take around three minutes of
iterative processing and substantial API cost per image, while Claude Code drops to
40.91\% on Grade-2. Models also
often fail silently, producing plausible but incorrect transcriptions
rather than abstaining, which is especially harmful for an assistive tool
\cite{bang-etal-2025-hallulens,
jackson2025aaomniscienceevaluatingcrossdomainknowledge}.

A compact model that reads Braille directly would avoid this overhead,
but training one would ordinarily require large amounts of labeled
real-world images, which are slow to collect and require Braille-literate
annotators. 
Earlier work has shown that multimodal models can learn Braille
transcription from synthetic images
\cite{karamolegkou2025evaluatingmultimodallanguagemodels}.
However, that work evaluates only on rendered images,
leaving open how well such models read photographed Braille.
We examine this transfer using BrailleBench, a benchmark of
real-world Braille images. We also test whether accuracy on synthetic
validation data predicts accuracy on real photographs. We train entirely on synthetic data: we generate a
corpus of 276k samples, rendered and degraded Braille image--text pairs, and
use it to adapt pretrained vision-language and OCR models. Without seeing
any real-world Braille images during training, our Qwen3.5-2B model reaches
86.98\% exact match on Grade-1, far above every standalone VLM and close
to Claude Code using a single model call, and 61.36\% exact match on Grade-2, 20.45 percentage points above Claude Code. In contrast, Qwen3.5-2B trained on the prior work's synthetic data reaches 96\% exact match on its synthetic validation set but only 3.26\% on Grade-1 BrailleBench, showing that synthetic validation scores do not predict real-world accuracy. The remaining errors
show that the synthetic-to-real transfer is not yet solved.

Our contributions are:
\begin{itemize}
    \item \textbf{BrailleBench}, a manually verified real-world benchmark
    of Grade-1 and Grade-2 Braille images, includes an evaluation and error
    analysis of proprietary, open-weight, OCR-specialized, and
    tool-assisted systems.

    \item \textbf{A synthetic Braille generation pipeline}, along with compact
 transcription models trained on it, outperforms all standalone VLMs
    on real-world Braille without using any real-world training images.

    \item \textbf{A synthetic-to-real analysis} shows that accuracy on
    synthetic validation data does not predict accuracy on real Braille, and
    that changing the training corpus substantially improves transfer while
    holding the model architecture and adaptation method fixed.
\end{itemize}

\section{Related Work}
\label{sec:related_work}

\paragraph{Multimodal models and visual text understanding.}
Modern vision-language models have evolved from early systems such as BLIP~\cite{li2022blip}, Flamingo~\cite{alayrac2022flamingo}, and LLaVA
\cite{liu2023visual}
to large-scale multimodal models trained on broad image--text corpora
\cite{beyer2024paligemma,schuhmann2022laion5bopenlargescaledataset,
li2024omnicorpus,zhu2023multimodal}. Despite strong performance on benchmarks such as VQA, ChartQA, OmniDocBench, OCRBench, and OCRBench v2 \cite{antol2015vqa,masry2022chartqa,ouyang2025omnidocbench,
Liu_2024,fu2025ocrbenchv2}, fine-grained text perception remains challenging.

\paragraph{OCR and synthetic supervision.}
Transformer-based and multimodal OCR systems such as TrOCR, Donut, Nougat, GOT, PaddleOCR-VL, and Chandra-OCR have substantially improved printed, handwritten, and document-level text recognition
\cite{li2023trocr,kim2022donut,blecher2023nougat,wei2024got,
cui2025paddleocrvl,paruchuri2026chandra}.
Synthetic supervision has also become increasingly important for OCR, with systems such as TrOCR, SynthDoG, and Nemotron-OCR demonstrating that large-scale generated image--text pairs can reduce dependence on manually labeled data
\cite{li2023trocr,kim2022donut,chesler2026ocr_synthetic_multilingual,
nemotronocrv2}.
Our work extends this direction to synthetic-to-real Braille transcription. 
\paragraph{Braille recognition.}
Optical Braille Recognition (OBR) has been studied since the early 1990s
\cite{mennens1993optical,Isayed2015ARO}. Early methods relied on hand-crafted
image processing, thresholding, de-skewing, and Braille-grid construction
\cite{Antonacopoulos2004ARB,babadi2011novel}, followed by machine-learning
approaches using SVMs, Haar-like features, HOG/LBP descriptors, and multi-stage
detection pipelines \cite{li2010optical,li2019effective}. Later neural methods
used cellular networks, autoencoders, CNNs, semantic segmentation, and object
detection for Braille dots, characters, and pages
\cite{Zhang2003CellularNN,Li2014ADL,Shimomura2017ConstructionOR,
li2018dsbi,li2020braunet,ovodov2021optical}. While these works established
strong Braille-specific recognition systems, they generally rely on specialized
architectures or constrained pipelines.
\citet{karamolegkou2025evaluatingmultimodallanguagemodels}
demonstrate Braille adaptation using synthetic images and LoRA~\cite{hu2022lora}.
Their transcription experiments train on 100k English WikiMatrix sentences
from the WMT 2024 shared-task data and evaluate on rendered FLORES-200 and
NTREX-128 images, so both training and evaluation use synthetic renders. We build on
this work by evaluating transfer to photographed Braille using
BrailleBench to assess whether synthetic-only adaptation transfers beyond rendered images.
\section{Methodology}
\label{sec:methodology}
This section describes BrailleBench, our synthetic data generation pipeline, the models we adapt, the prior synthetic data we use to study synthetic-to-real transfer, and our evaluation setup.
\subsection{BrailleBench}
\label{subsec:data_collection}

We construct \textbf{BrailleBench}, a real-world benchmark for evaluating visual Braille transcription. We collect Braille images from publicly available Internet sources and books, spanning a range of text lengths, materials, illumination conditions, and viewing angles. Each image is manually inspected and transcribed by two annotators, and the annotations are subsequently cross-reviewed by the annotators. After removing ambiguous, unreliable, and duplicate samples, the final benchmark contains \textbf{259 images} of which 215 are Grade-1 and 44 Grade-2. English Braille is written in two forms. \emph{Grade-1} (uncontracted) Braille generally represents letters individually and uses additional indicators for features such as capitalization and numbers; it does not have a universal one-to-one mapping between printed characters and Braille cells \cite{iceb2024ueb}. \emph{Grade-2} (contracted) Braille, the form used in most books and signage, additionally uses contractions in which a single cell or a short sequence of cells stands for a whole word or a common letter group; for example, the letter \textit{c} written alone denotes the word \textit{can}. Because the same cell can be read differently depending on its position and surroundings, recognizing Grade-2 Braille requires resolving contractions from context as well as perceiving the dots. Appendix~\ref{app:benchmark} gives per-grade statistics of text length. 

\subsection{Synthetic Training Corpus}
\label{subsec:synthetic_corpus}

\begin{figure*}[t]
    \centering
    \includegraphics[width=\textwidth]{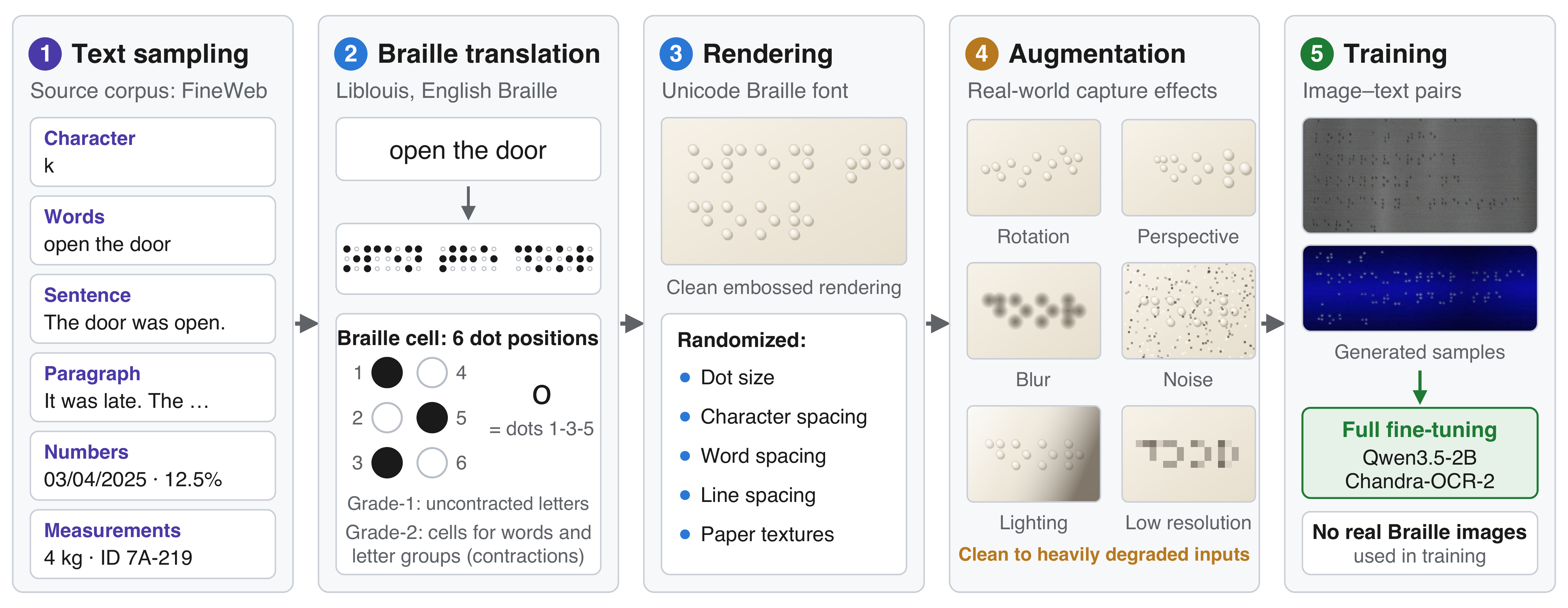}
    \caption{Synthetic training data generation pipeline. Text sampled from
    FineWeb at multiple granularities is translated to Braille with Liblouis,
    rendered as embossed Braille with randomized dot geometry, spacing, and
    paper textures, and degraded with augmentations that simulate real-world
    capture conditions. The resulting image--text pairs are used to fine-tune
    Qwen3.5-2B and Chandra-OCR-2 without any real-world Braille images.}
    \label{fig:synthetic_pipeline}
\end{figure*}

Because large-scale labeled real-world Braille imagery is difficult to obtain, we generate the entire training corpus programmatically. No real-world Braille images are used for training. Figure~\ref{fig:synthetic_pipeline} summarizes the generation pipeline. We generate \textbf{around 276k samples} by sampling English text and converting it to its Braille equivalent with Liblouis \cite{egli2009liblouis} to render images. We use FineWeb \cite{penedo2024the} as the primary source text corpus because of its greater diversity in vocabulary, style, and topic. 
Text is sampled at multiple granularities, including single words, short phrases, sentences, and paragraphs, as well as additional structured numerical expressions such as dates, percentages, measurements, and identifiers. 
For image augmentation, we simulate embossed Braille under randomized illumination and viewing conditions using a Unicode-font-based renderer with paper textures and geometric deformation. Dot size, character spacing, word spacing, and line spacing are randomized to increase geometric diversity. To improve robustness to real-world capture conditions, we apply random rotation, perspective distortion, blur, noise, lighting variation, and resolution degradation, following synthetic OCR augmentation strategies, such as those of \citet{chesler2026ocr_synthetic_multilingual}. These transformations produce examples ranging from clean rendered Braille to low-contrast and heavily degraded inputs. Layout parameters such as cell pitch, dot size, and spacing are matched to measurements from real Braille photographs that are disjoint from BrailleBench. About half of the corpus is Grade-2, and a quarter of the texts have scrambled letters, so that the model must read every cell instead of guessing words from context. Appendix~\ref{app:synthetic} lists all generation parameters and shows example renders.

\paragraph{Corpus variants.}
Our main training corpus contains 276{,}717 samples (``276k''). Its texts are short, averaging 39 characters (median four words), which is close to the short texts typical of BrailleBench (22 characters on average). To test whether more and longer synthetic data helps, we also generate a 500k-sample corpus (``500k'') with the same pipeline and text source, but with longer text windows: 148 characters and 25 words on average, up to 300 characters. Table~\ref{tab:training_corpus_profiles} compares the two corpora. Unless stated otherwise, ``our model'' refers to the model trained on the 276k corpus. The 500k corpus is used only in the analysis of Section~\ref{subsec:syn_to_real} (details in Appendix~\ref{app:wmt_eval}).

\subsection{Model Adaptation}
\label{subsec:training}

We investigate whether Braille transcription can be learned entirely from synthetic supervision. We adapt Qwen3.5-2B~\cite{qwen35blog} and the 5B Chandra-OCR-2~\cite{paruchuri2026chandra} vision-language models.

Each model is trained on the same set of synthetically generated data with an image-to-text transcription objective. We adopt full fine-tuning of all model parameters. The resulting models are evaluated directly on BrailleBench, allowing us to measure synthetic-to-real transfer without training on real-world benchmark images. The evaluation setup is described in Section~\ref{subsec:evaluation}, and training hyperparameters are listed in Appendix~\ref{sec:training_hyperparameters}.

\subsection{Synthetic-to-Real Comparison}
\label{subsec:syn_real_setup}

To test whether accuracy on synthetic images predicts accuracy on real photographs, we use the synthetic Braille dataset released by prior work \cite{karamolegkou2025evaluatingmultimodallanguagemodels}: 100k rendered and augmented images of English sentences from the WikiMatrix corpus of the WMT 2024 shared task \cite{wmt-2024-1}. WMT serves only as a source of English sentences; no translation is involved. We split this \emph{WMT-Braille} data into \emph{WMT-Train} (99k images) and a held-out \emph{WMT validation} set (1k images). Following the prior work, we train Llama-3.2-11B with LoRA on WMT-Train, and additionally Qwen3.5-2B with full fine-tuning. We evaluate both on WMT validation, which is in-distribution for them, and on BrailleBench. Inference and scoring are identical for all evaluation sets.
\subsection{Evaluation}
\label{subsec:evaluation}

We evaluate three groups of systems on BrailleBench: (i) zero-shot general-purpose VLMs, (ii) tool-assisted agentic harnesses, and (iii) Braille-specific systems, namely an existing optical Braille recognition (OBR) tool, a reproduction of prior synthetic adaptation, and our synthetic-adapted models.

\subsubsection{Baselines}
\label{subsec:baselines}

\paragraph{General-purpose VLMs.}
We evaluate proprietary models GPT-5.5~\cite{openai2026gpt55}, Claude Opus 4.8 \cite{anthropic2026opus48}, Gemini 3.5 Flash \cite{googledeepmind2026gemini35flash}, GPT-5.6 Sol High \cite{openai2026gpt5.6}, and GPT-6.1 Sol \cite{openai2026gpt61sol}, along with the open-weight Kimi K2.6 \cite{moonshotai2026kimik26} and Gemma 4 31B-IT \cite{gemmateam2026gemma4technicalreport}. All are evaluated zero-shot using the prompt:
\texttt{Decode the Braille text shown in the image and output the transcription.}
They receive no Braille-specific fine-tuning or in-context examples, and the final transcription is extracted from conversational output before scoring.

\paragraph{Tool-assisted inference.}
We evaluate GPT-5.6 Sol in the Codex harness and Claude Opus 5.5 in Claude Code. Both can iteratively crop, enlarge, and filter the image with external tools, at about 180 seconds per image, so we report them separately from direct inference.

\paragraph{Existing OBR systems.}

We evaluate the Angelina Braille Reader \cite{ovodov2021optical} and reproduce the synthetic adaptation of \citet{karamolegkou2025evaluatingmultimodallanguagemodels}: Llama-3.2-11B-Vision-Instruct fine-tuned with LoRA on WMT-Train (Section~\ref{subsec:syn_real_setup}).
\subsubsection{Metrics}
\label{subsec:metrics}

Predictions are compared with the manually verified references at the character and word level. We report character and word error rates (CER, WER; edit distance normalized by reference length, lower is better), exact match (the whole transcription is correct), chrF (character $n$-gram overlap), normalized match (exact match after Unicode normalization, lowercasing, and whitespace normalization; Appendix~\ref{app:evaluation}), Word F1 (token overlap), and edit similarity. Exact match is the strictest measure, since a single wrong character makes the transcription incorrect.

\section{Experiments and Results}
\label{sec:results}

Table~\ref{tab:braille-all-models} reports results separately for the 215 Grade-1 and 44 Grade-2 images. CER and WER above 1 come from long hallucinated outputs.

\subsection{Main Results}

\begin{table*}[!t]
\centering
\small
\setlength{\tabcolsep}{2.5pt}
\renewcommand{\arraystretch}{1.05}
\begin{tabularx}{\textwidth}{
>{\raggedright\arraybackslash}p{0.24\textwidth}
>{\centering\arraybackslash}p{0.05\textwidth}
*{7}{>{\centering\arraybackslash}X}
}
\toprule
Model & Grade & CER $\downarrow$ & WER $\downarrow$ &
\shortstack{Exact\\Match\\(\%) $\uparrow$} & chrF $\uparrow$ &
\shortstack{Norm.\\Match\\(\%) $\uparrow$} &
\shortstack{Word F1\\(\%) $\uparrow$} &
\shortstack{Edit\\Similarity\\(\%) $\uparrow$} \\

\midrule
\multicolumn{9}{c}{\textbf{Standalone General-Purpose VLMs}} \\
\midrule
GPT-5.5
& G1 & 1.147 & 1.176 & 0.00 & 2.10 & 2.79 & 5.55 & 19.32 \\
& G2 & 0.776 & 0.882 & 0.00 & 11.13 & 0.00 & 6.99 & 23.04 \\[2pt]

Kimi K2.6
& G1 & 1.276 & 1.516 & 0.00 & 5.68 & 0.00 & 0.59 & 12.82 \\
& G2 & 0.941 & 1.077 & 0.00 & 5.79 & 0.00 & 2.70 & 13.98 \\[2pt]

Claude Opus 4.8
& G1 & 0.934 & 1.157 & 0.93 & 8.76 & 1.40 & 3.86 & 18.12 \\
& G2 & 0.755 & 0.889 & 0.00 & 13.97 & 2.27 & 11.75 & 25.04 \\[2pt]

Gemini 3.5 Flash
& G1 & 0.773 & 0.973 & 2.79 & 11.94 & 4.65 & 10.15 & 26.39 \\
& G2 & 0.712 & 0.853 & 0.00 & 13.13 & 2.27 & 14.63 & 28.41 \\[2pt]

Gemma 4 31B-IT
& G1 & 7.290 & 18.170 & 0.47 & 5.67 & 0.93 & 2.20 & 15.68 \\
& G2 & 0.791 & 0.967 & 0.00 & 9.90 & 0.00 & 7.78 & 22.02 \\[2pt]

GPT-5.6 Sol High
& G1 & 0.900 & 1.091 & 11.16 & 17.43 & 15.81 & 18.09 & 33.66 \\
& G2 & 0.808 & 0.858 & 4.55 & 20.78 & 4.55 & 18.63 & 29.48 \\[2pt]

GPT-6.1 Sol
& G1 & 0.374 & 0.412 & 53.95 & 67.32 & 56.28 & 65.87 & 70.82 \\
& G2 & 0.944 & 0.944 & 11.36 & 20.50 & 11.36 & 22.29 & 23.20 \\[2pt]

\midrule
\multicolumn{9}{c}{\textbf{Tool-Assisted Harnesses}} \\
\midrule
Claude Code (Opus 5.5 High)
& G1 & \textbf{0.019} & \textbf{0.058} & \textbf{88.37} & \textbf{97.00} & \textbf{92.56} & \textbf{95.78} & \textbf{98.74} \\
& G2 & 0.252 & 0.458 & 40.91 & 79.11 & 40.91 & 69.34 & 81.41 \\[2pt]

Codex (GPT-5.6 Sol)
& G1 & 0.244 & 0.335 & 62.79 & 75.59 & 64.19 & 74.24 & 79.96 \\
& G2 & 0.994 & 1.059 & 0.00 & 7.61 & 0.00 & 0.00 & 12.95 \\[2pt]

\midrule
\multicolumn{9}{c}{\textbf{Existing Braille Systems}} \\
\midrule
Angelina Braille Reader~\cite{ovodov2021optical}
& G1 & 0.565 & 0.816 & 3.26 & 26.01 & 4.19 & 19.67 & 37.39 \\
& G2 & 0.512 & 0.658 & 0.00 & 32.36 & 0.00 & 30.80 & 47.54 \\[2pt]

Llama-3.2-11B (WMT-Train)
& G1 & 21.961 & 34.197 & 0.47 & 7.88 & 0.47 & 4.48 & 14.85 \\
& G2 & 1.140 & 1.468 & 0.00 & 8.13 & 0.00 & 4.82 & 18.60 \\[2pt]
\midrule
\multicolumn{9}{c}{\textbf{Synthetic-Adapted Models (Ours)}} \\
\midrule
Qwen3.5-2B (Ours)
& G1 & 0.025 & 0.083 & 86.98 & 95.85 & 87.91 & 93.11 & 97.48 \\
& G2 & \textbf{0.035} & \textbf{0.069} & \textbf{61.36} & \textbf{90.68} & \textbf{68.18} & 91.82 & \textbf{94.78} \\[2pt]

Chandra-OCR-2 (Ours)
& G1 & 0.034 & 0.093 & 85.12 & 94.71 & 86.98 & 93.13 & 96.82 \\
& G2 & 0.055 & 0.093 & 54.55 & 89.96 & 59.09 & \textbf{92.38} & 94.04 \\[2pt]

\bottomrule
\end{tabularx}
\caption{
Performance comparison of standalone general-purpose VLMs,
tool-assisted harnesses, existing Braille systems (an OBR tool and our
reproduction of prior synthetic adaptation; \citealp{karamolegkou2025evaluatingmultimodallanguagemodels}),
and our synthetic-adapted models on BrailleBench, reported separately
for the 215 Grade-1 (G1) and 44 Grade-2 (G2) images. Lower CER and WER
indicate better performance, while higher values are preferred for all
other metrics. Bold marks the best value per grade.
}
\label{tab:braille-all-models}
\end{table*}

\paragraph{Standalone VLMs.}
Most standalone VLMs cannot read Braille directly: on Grade-1, all except the GPT Sol models stay below 3\% exact match, and Gemma 4 31B-IT's long hallucinated outputs push its CER to 7.29. GPT-6.1 Sol is by far the strongest standalone model and leads every Grade-1 metric (53.95\% exact match, CER 0.374, WER 0.412), a large step over GPT-5.6 Sol High (11.16\%). Grade-2 is harder for all of them: only GPT-5.6 Sol High (4.55\%) and GPT-6.1 Sol (11.36\%) produce any exact match, and GPT-6.1 Sol's CER rises to 0.944.

\paragraph{Tool-assisted inference.}

Both harnesses outperform every standalone VLM on Grade-1: Codex reaches 62.79\% exact match (vs.\ 11.16\% for GPT-5.6 Sol without tools), and Claude Code is the strongest Grade-1 system on every metric (88.37\%, CER 0.019). On Grade-2, Claude Code drops to 40.91\% (CER 0.252), and Codex fails entirely (0\%, CER 0.994), doing worse than GPT-5.6 Sol without tools. Enlarging and cropping likely help the model see the dots but not resolve context-dependent Grade-2 contractions. The Claude Code logs show image-processing tools producing views from which the model reads the dots. Codex also executes generated code, so its tool use is not restricted to cropping, enlargement, and filtering (Appendix~\ref{app:harness}). The reported average processing time is about three minutes per image, with an API-equivalent cost of \$0.51--0.58 per image.

\begin{figure}[t]
    \centering
    \includegraphics[width=1.0\columnwidth]{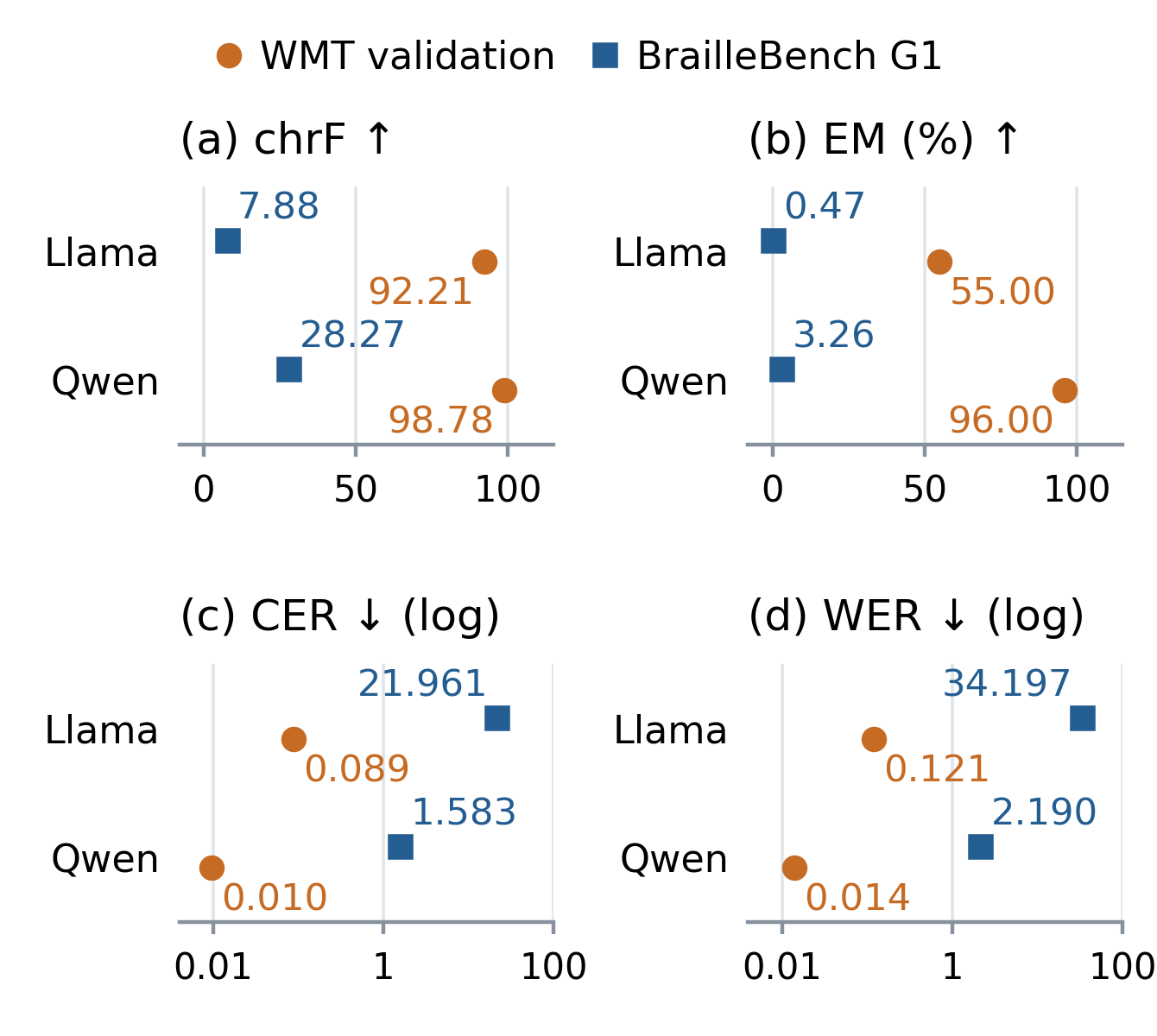}
    \caption{Models trained only on WMT-Train (Llama-3.2-11B with LoRA; Qwen3.5-2B with full fine-tuning), evaluated on the in-distribution WMT validation set and on Grade-1 BrailleBench (both models reach 0\% exact match on Grade-2). Both models score highly on synthetic validation but fail on real photographs. CER and WER use logarithmic axes; values above 1 come from long hallucinated outputs.}
    \label{fig:wmt_transfer}
\end{figure}

\paragraph{Synthetic-adapted models.}
Both of our models outperform all standalone VLMs, Codex, and the existing Braille systems on every Grade-1 metric; Qwen3.5-2B (86.98\%, CER 0.025) is only 1.39 points below Claude Code, and Chandra-OCR-2 follows closely (85.12\%, CER 0.034). The prior-work recipe, Llama-3.2-11B trained on WMT-Train, reaches only 0.47\% (Section~\ref{subsec:syn_to_real}). On Grade-2 they achieve the best scores of all systems (on 44 images): Qwen3.5-2B reaches 61.36\% exact match and a CER of 0.035, compared with 40.91\% and 0.252 for Claude Code, and Chandra-OCR-2 has the best Grade-2 Word F1. Qwen's CER stays low on both grades, while its exact match drops on Grade-2. Grade-2 references are about six times longer (71.7 vs.\ 12.1 characters on average; Table~\ref{tab:benchmark-statistics}), which increases the opportunity for a transcription to contain an error. Text length and contraction complexity are confounded in this comparison. Overall, Claude Code leads on Grade-1 and our models on Grade-2. Our models need a single call to a 2B or 5B model per image, without a tool loop, instead of about three minutes of agentic processing and \$0.51--0.58 in API usage, and use no real training images. The remaining errors, especially on long Grade-2 texts, show that synthetic-to-real transfer is not yet solved. Section~\ref{sec:error_analysis} analyzes how each system fails, and Appendix~\ref{app:qualitative} adds results on an out-of-benchmark Braille book.

\subsection{Transfer from Synthetic to Real Braille}
\label{subsec:syn_to_real}

\paragraph{Training on prior synthetic data.}
Can models trained on the prior work's synthetic data (WMT-Train; Section~\ref{subsec:syn_real_setup}) read photographed Braille? Figure~\ref{fig:wmt_transfer} shows they cannot. Llama-3.2-11B with LoRA reaches 55.00\% exact match on WMT validation but 0.47\% on Grade-1 BrailleBench, where it often produces long hallucinated outputs (CER 21.96). Qwen3.5-2B with full fine-tuning reaches 96.00\% on WMT validation but only 3.26\% on Grade-1 and 0\% on Grade-2, and its normalized Grade-1 match of 17.67\% shows that formatting explains little of the gap. Near-perfect synthetic validation accuracy thus says little about reading real Braille, which is why we evaluate on BrailleBench.

\paragraph{Is it the data or the model?}
The prior-work recipe differs from ours in both training data and adaptation method (LoRA instead of full fine-tuning). To separate the two, we train the same Llama-3.2-11B with LoRA on our 276k corpus at ranks 8, 16, and 32 (Table~\ref{tab:lora_rank_main}). Every rank reaches 72--76\% exact match on BrailleBench, whereas the WMT-trained model gets one of 259 images right. Higher ranks mainly curb over-long outputs: rank 8 produces almost twice as many words as the references, raising its CER to 0.42, versus about 0.076 for ranks 16 and 32. Changing the training corpus substantially improves transfer with the same architecture and LoRA adaptation method (training curves in Appendix~\ref{app:lora_rank}).

\paragraph{Does more, longer synthetic data help?}
Not in this comparison. Training Qwen3.5-2B on the 500k corpus, which has about four times longer texts, lowers BrailleBench exact match (both grades) from 82.63\% to 72.97\%, although it raises exact match on the longer WMT validation sentences from 41.70\% to 63.10\%. Selected on synthetic validation alone, the worse model would be chosen. Continual training of the 500k model on 15k short samples recovers most of the gap (81.85\%), which is consistent with a role for text length; corpus size and image dimensions also differ, so these factors are not isolated (Appendix~\ref{app:wmt_eval} and~\ref{app:continual_training}).

\begin{table}[t]
\centering
\small
\setlength{\tabcolsep}{3pt}
\renewcommand{\arraystretch}{1.1}
\begin{tabular}{llrrrr}
\toprule
Training data & Rank & chrF $\uparrow$ & EM (\%) $\uparrow$ & CER $\downarrow$ & WER $\downarrow$ \\
\midrule
Ours (276k) & 8  & 88.28 & 72.59 & 0.4196 & 1.0704 \\
Ours (276k) & 16 & 88.57 & 72.20 & \textbf{0.0754} & 0.1444 \\
Ours (276k) & 32 & \textbf{90.10} & \textbf{76.06} & 0.0759 & \textbf{0.1261} \\
\bottomrule
\end{tabular}
\caption{Llama-3.2-11B-Vision-Instruct trained with LoRA on our 276k corpus, evaluated on BrailleBench (both grades, $n=259$). The same model trained on WMT-Train transcribes only one of the 259 images exactly (Table~\ref{tab:braille-all-models}).}
\label{tab:lora_rank_main}
\end{table}

\section{Error Analysis}
\label{sec:error_analysis}

\begin{figure*}[t]
    \centering
    \includegraphics[width=\textwidth]{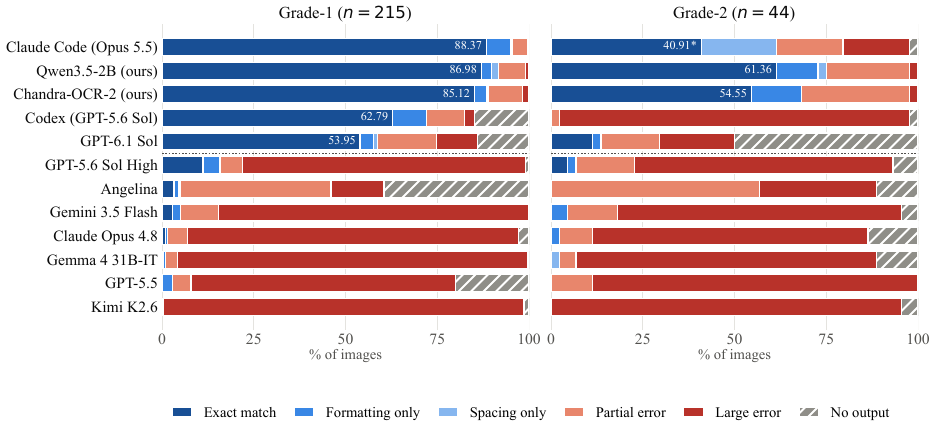}
    \caption{Prediction outcomes by grade. Each image falls into exactly one category. \emph{Formatting only}: differences that disappear after normalization; \emph{spacing only}: only word spacing differs; \emph{partial error}: at most half of the normalized reference characters need editing; \emph{large error}: more than half; \emph{no output}: empty outputs and refusals. Numbers on bars give exact match (\%). Rows above the dotted line are the harnesses, our models, and the strongest standalone model. Exact counts are in Appendix~\ref{app:error_counts}. For Claude Code Grade-2 (*), categories are recomputed from the predictions using the normalized-match transform in Table~\ref{tab:braille-all-models}, retaining punctuation. Other rows retain the supplied counts under the original looser formatting normalization; their error-type splits are not directly comparable to this recomputed row. Exact match remains raw string equality for all rows.}
    \label{fig:error_breakdown}
\end{figure*}

Figure~\ref{fig:error_breakdown} groups prediction outcomes into error categories; different images that share the same text are kept as separate samples. Our models almost never produce large errors or no output (at most 4 of 215 Grade-1 and 1 of 44 Grade-2 images); most of their mistakes are partial or formatting-only. Claude Code behaves similarly on Grade-1. Its recomputed Grade-2 breakdown contains 18 exact matches, 9 spacing-only errors, 8 partial errors, 8 large errors, and 1 empty output, using the protocol specified in Appendix~\ref{app:error_counts}. Codex returns nothing for 32 Grade-1 images and makes a large error on 42 of the 44 Grade-2 images, and GPT-6.1 Sol returns nothing for 30 Grade-1 and 22 Grade-2 images. The other standalone VLMs make large errors on 155--211 of the 215 Grade-1 images. Many of these large errors are fluent text, such as the repeated guesses described below, rather than visible noise, so they fail silently: a user who cannot read Braille cannot tell them apart from a correct transcription, whereas an empty output is at least noticeable. Angelina, a dedicated OBR system, fails differently: most of its errors are partial (88) or empty (85) on Grade-1.

Inspecting the predictions themselves, we see four recurring patterns. Unless stated otherwise, counts refer to the 215 Grade-1 images.

\paragraph{Letter, word, and spacing errors.}
Qwen and Chandra mostly make small changes to the text. Examples include Qwen reading \texttt{staff only} as \texttt{stabf only}, and Claude Code joining \texttt{when Oscar} into \texttt{whenOscar}. In the recomputed Claude Code Grade-2 row, 9 of 44 predictions differ only in spacing under the stated normalization. Small errors can still change meaning: Qwen replaces \texttt{like to} with \texttt{cannot} in one passage.

\paragraph{Hallucinations.}
On Grade-1, Claude Opus 4.8 repeatedly outputs unrelated words such as \texttt{can} (14 images) and \texttt{Braille} (8). GPT-5.6 Sol High outputs \texttt{hello world} on eight unrelated images. Such repeated wrong guesses (excluding correct readings) occur in 76 predictions for Claude Opus 4.8, 44 for GPT-5.6 Sol High, and only 11 for GPT-6.1 Sol.

\paragraph{Missing text.}
GPT-6.1 Sol often leaves out words or produces no text. It outputs blank results for 30 out of 215 Grade-1 images and 22 out of 44 Grade-2 images. For example, it returns only \texttt{we} for \texttt{Lucy saw it when Oscar opened his book.}

\paragraph{Longer text.}
Using normalized match, GPT-5.6 Sol High's Grade-1 accuracy falls from 27.3\% on one-word targets to 6.0\% on multiword targets; GPT-6.1 Sol falls from 73.7\% to 41.4\%. Among the 20 Grade-2 targets with at least seven words, Qwen correctly reads 10, Claude Code reads two, and GPT-6.1 Sol reads none. Longer text is harder, although length and grade effects overlap because most long samples are Grade-2. Appendix~\ref{app:harness_failures} shows qualitative examples of fluent but wrong outputs from the tool-assisted harnesses.

\section{Conclusion}
\label{sec:conclusion}

We introduced \textbf{BrailleBench}, a real-world benchmark for visual Braille transcription. Standalone general-purpose VLMs remain weak at reading Braille, while tool-assisted harnesses do much better at a high cost. Models trained only on our synthetic corpus outperform every standalone VLM on Grade-1, come close to the best harness, and achieve the best Grade-2 scores of all systems, with a single model call per image. In our experiments, high synthetic validation accuracy does not guarantee accurate transcription of real photographs, and changing the training corpus substantially improves transfer with the model architecture fixed. Real-image benchmarks are therefore needed to measure progress. This work lays a foundation for systems that support Braille education, communication between Braille and non-Braille readers, and document digitization. More broadly, accessibility-relevant capabilities should be evaluated explicitly rather than assumed to follow from strong general multimodal performance.

\section*{Limitations}
\label{sec:limitations}

BrailleBench is intentionally focused in scope. The benchmark contains 259 manually curated real-world Braille images, and its size and collection process may introduce sampling bias. Because the images originate from publicly available sources, we cannot rule out possible overlap with the pretraining data of some foundation models. Our evaluation uses fixed model versions and model APIs, along with a standardized prompt, without any per-model prompt optimization or confidence calibration. Although the layout measurements used for generation come from photographs disjoint from BrailleBench, inspecting BrailleBench images informed our choice of augmentations. In addition, the current study focuses on English Braille only and does not cover multilingual Braille or the full range of real-world capture conditions. Although synthetic adaptation substantially improves performance, the remaining errors across all trained models on BrailleBench indicate that synthetic-to-real generalization is still lacking. In particular, Grade-2 Braille
introduces additional complexity through contractions, whose interpretation can depend on linguistic context, word boundaries, and surrounding symbols. Such ambiguity makes recognition more challenging, especially for short or isolated text where contextual information is limited.

Future work will systematically investigate these ambiguities, expand the evaluation of multilingual Braille across diverse real-world conditions, and explore context-aware recognition methods.

The Grade-2 subset contains only 44 images, so a small number of predictions can change its accuracy substantially. Reported differences are descriptive; we do not establish paired statistical significance from the aggregate scores. Shared sources and repeated target strings may also introduce dependencies between images.

\section*{Ethics Statement}

This research focuses on Braille transcription using Optical Braille Recognition and its potential applications in education and everyday communication. Informal conversations with schools, Braille training centers, and community members highlighted difficulties with early Braille learning and interpreting teaching materials, particularly for teachers who do not read Braille fluently. Transcription tools could help teachers review student work and assist families in understanding Braille text. We treat this feedback as qualitative motivation to investigate transcription support rather than representative evidence of national literacy rates or teacher shortages. Our recognition experiments do not establish improvements in learning outcomes or demonstrate that automated transcription can replace specialist teaching.

Responsible assistive technology requires evaluation under realistic conditions. We therefore examine recognition on photographed Braille alongside synthetic images, rather than assuming that performance on rendered images establishes real-world reliability. Nevertheless, our models still produce errors, including substitutions that change meaning. Incorrect transcriptions could mislead learners or distort the information in the labels and instructions. Future applications should communicate uncertainty, support verification by Braille-literate users, and avoid presenting uncertain predictions as authoritative. Cell-level explanations and educational feedback would require separate human verification and validation.
Our benchmark focuses on English Braille and cannot establish performance across all languages, Braille conventions, or capture conditions. Future development should involve blind and low-vision users and Braille educators in design and evaluation, with an eye on accessibility and affordability. Braille photographs and their transcriptions may contain personal information; responsible collection, processing, and release require privacy protections and appropriate source permissions. We encourage further research that expands access to Braille learning while preserving users' agency and independent literacy.
\bibliography{references}

\clearpage
\appendix
\renewcommand{\topfraction}{0.9}
\renewcommand{\bottomfraction}{0.8}
\renewcommand{\textfraction}{0.07}
\renewcommand{\floatpagefraction}{0.8}
\renewcommand{\dbltopfraction}{0.9}
\renewcommand{\dblfloatpagefraction}{0.8}
\setcounter{topnumber}{4}
\setcounter{bottomnumber}{4}
\setcounter{totalnumber}{8}
\setcounter{dbltopnumber}{4}

\section{BrailleBench Details}
\label{app:benchmark}

BrailleBench contains 259 real-world Braille images, of which 215 are
Grade-1 (uncontracted) and 44 are Grade-2 (contracted). Images were
collected from publicly available Internet sources and from Braille books,
and cover embossed paper, printed Braille, signage, and labels photographed
under varied illumination and viewing angles. Each image was transcribed
independently by two annotators, and the transcriptions were then
cross-reviewed; ambiguous, unreliable, and duplicate images were removed.
Different images that share the same text are retained.

\begin{table}[!htbp]
\centering
\small
\setlength{\tabcolsep}{4pt}
\renewcommand{\arraystretch}{1.12}
\begin{tabular}{@{}lrr@{}}
\toprule
Statistic & Grade 1 & Grade 2 \\
\midrule
Images & 215 & 44 \\
\midrule
Characters: mean & 12.06 & 71.66 \\
Characters: median & 9 & 34.5 \\
Characters: maximum & 61 & 368 \\
\midrule
Words: mean & 2.25 & 13.89 \\
Words: median & 2 & 6 \\
Words: maximum & 10 & 75 \\
\midrule
Single-word targets & 99 & 5 \\
Multiword targets & 116 & 39 \\
Targets with 2--6 words & 112 & 19 \\
Targets with $\geq 7$ words & 4 & 20 \\
\bottomrule
\end{tabular}
\caption{BrailleBench statistics by grade. Character
and word lengths are measured per reference.
Characters include spaces and punctuation; words
are whitespace-delimited. Multi-line image counts
are not annotated.}
\label{tab:benchmark-statistics}
\end{table}

\section{Synthetic Data Details}
\label{app:synthetic}

\subsection{Overview}
\label{app:synthetic_params}
\label{app:datagen}
Text windows are sampled from FineWeb \cite{penedo2024the} (\texttt{sample-10BT}; one add-on from CC-MAIN-2013-20),
supplemented with structured numerical expressions such as dates, percentages, measurements, and identifiers, and
translated to Unicode Braille with Liblouis \cite{egli2009liblouis}. The Braille is rendered as embossed dots with a
height-map shading model, degraded by the augmentations of Table~\ref{tab:aug}, and finally cropped and rescaled to a
target cell pitch drawn from measurements on real photographs (Table~\ref{tab:layout}). Every geometric step also
transforms the word/line boxes, so each image carries exact box labels.
The corpus is assembled from several generation runs (Table~\ref{tab:sources}). \emph{Real-matched} runs draw the layout
(cell pitch, dot size, spacing, margins, line count) from about 3k measurement records of real Braille photographs, disjoint from BrailleBench
(Table~\ref{tab:layout}). \emph{Add-on} runs target cases the base runs cover poorly: rare rendering styles such as flat-printed
``ghost'' cells, hollow dots, tape strips and close-ups, wider spacing ranges, and texts with \emph{scrambled} letters. Scrambled texts
have their letters randomly shuffled, so the model cannot guess words from language context and has to read every cell.
All statistics in this section describe the main 276k corpus (276{,}717 images). The 500k corpus used in
Appendix~\ref{app:wmt_eval} is produced by the same pipeline with longer text windows
(Table~\ref{tab:training_corpus_profiles}).

\subsection{Braille Translation and Corpus Composition}
Text is translated with Liblouis~3.38.0 using the table chain \texttt{braille-patterns.cti} +
\texttt{en-ueb-g1.ctb} for Grade~1 (uncontracted) and \texttt{braille-patterns.cti} + \texttt{en-ueb-g2.ctb}
for Grade~2 (contracted), i.e.\ Unified English Braille (UEB). All 276{,}717 samples use these tables.
The corpus is 47.4\% Grade~1 and 52.6\% Grade~2 (Table~\ref{tab:sources}).
The models therefore see contracted Braille during training. On a random sample of 20{,}000 Grade-2 training texts, the Grade-2 translation differs
from the Grade-1 translation in 90.3\% of the samples and uses on average
19.9\% fewer cells (cells$_{G2}$/cells$_{G1}$: mean 0.80,
5th percentile 0.62); this includes scrambled texts, which contract less than natural words.
These tables implement UEB, not the older US code (EBAE). Real Grade-2 images transcribed under EBAE may use contractions that differ from those seen in training.

\begin{table*}[!tp]\centering\small

\begin{tabular}{lrrrr}\toprule
Run & Images & G1 / G2 (\%) & Words (med.) & Scr.\ (\%) \\\midrule
Real-matched base (FineWeb sample-10BT) & 151{,}096 & 44.6 / 55.4 & 4 & 17.7 \\
Real-matched base (CC-MAIN-2013-20) & 20{,}000 & 43.2 / 56.8 & 4 & 18.3 \\
Add-on: ghost/hollow/strip/close-up, dynamic spacing & 65{,}857 & 38.4 / 61.6 & 2 & 40.1 \\
Add-on: 1--5 words, all scrambled & 9{,}767 & 50.3 / 49.7 & 3 & 99.2 \\
Add-on: 6--25 words, 50\% scrambled & 9{,}997 & 50.0 / 50.0 & 15 & 49.9 \\
Earlier generator (sample of 20k) & 20{,}000 & 100.0 / 0.0 & 3 & 0.0 \\
\midrule
Total & 276{,}717 & 47.4 / 52.6 & 4 & 25.8 \\\bottomrule
\end{tabular}
\caption{Composition of the training corpus by generation run. Grade columns are percentages (G1 / G2); ``words'' is the median
number of words per text; ``scr.'' is the percentage of texts whose letters were scrambled. The earlier-generator part is Grade~1 only.}
\label{tab:sources}
\end{table*}

\paragraph{Earlier-generator part.}
20{,}000 images (7.2\%) come from an earlier version of the generator (Grade~1 only, same Liblouis table):
character spacing $\times$0.91--1.17, word spacing $\times$0.89--1.51, dot scale 0.93--1.38 (realised p5--p95),
29.7\% down-scaled to a random low resolution; no real-photo matching of scale or margins.

\subsection{Text Granularity}
Labels are random windows of web text, not linguistically segmented units, so the categories below are defined by length:
character = one non-space character, word = one token, phrase = 2--5 tokens, sentence-length = 6--25 tokens,
paragraph-length = 26 or more tokens. Numeric expressions are a separate, overlapping flag. Of the samples, 31.5\% contain
sentence-final punctuation and 5.1\% contain two or more sentences; 13.3\% contain a digit and 0.4\% consist
only of numeric tokens. Layout: 53.2\% of the images show one text line, 22.4\% two or three lines and 24.4\% four or more.
Single characters are almost absent (Table~\ref{tab:gran}); a dedicated character-level split is not part of this corpus.

\begin{table}[!htbp]\centering\small

\begin{tabular}{lr}\toprule
Granularity & \% of images \\\midrule
character & 0.0 \\
word & 20.6 \\
phrase (2-5 words) & 43.6 \\
sentence-length (6-25 words) & 34.1 \\
paragraph-length (26+ words) & 1.7 \\
\midrule
\emph{numeric expression (overlapping)} & 13.3 \\
\emph{letters scrambled (overlapping)} & 25.8 \\\bottomrule
\end{tabular}
\caption{Text granularity of the training labels (\% of images). Word-count bins: 1: 20.6, 2: 17.0, 3: 11.4,
4--5: 15.2, 6--9: 9.5, 10--15: 9.8, 16--25: 14.7, 26+: 1.7 (\%).}
\label{tab:gran}
\end{table}

\subsection{Rendering and Augmentation}
Braille cells are drawn with the \texttt{FreeMono} glyphs at a base size of 140\,px (cell advance 84\,px, dot pitch 26.5\,px, dot diameter
18\,px) and then rescaled; all pixel values in Table~\ref{tab:aug} refer to this base render, before the final resize.
Dots are drawn individually (so spacing is free), jittered, blurred into a height map and lit with a random directional light
(Lambertian shading; surface-grain noise $\mathcal N(1,0.08)$). Probabilities are per image and apply to the bulk of the corpus
(real-matched base runs; the add-on runs change the entries marked $\dagger$).

\begin{figure*}[!tp]\centering
\includegraphics[width=0.78\textwidth]{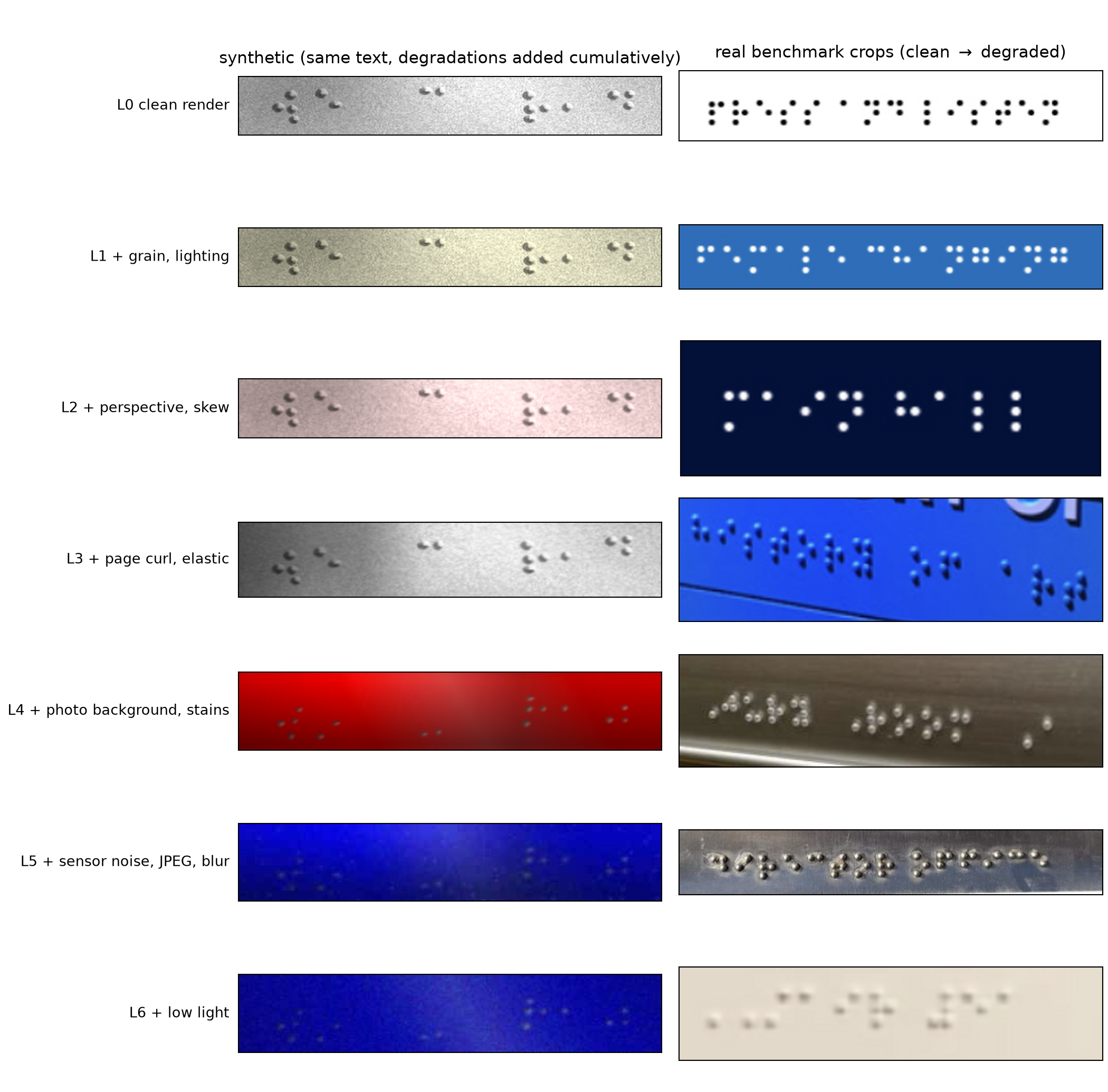}
\caption{Left: one text (\texttt{we can read}, Grade~2) rendered with degradations added cumulatively, from a clean render (L0) to heavy
degradation (L6); layout is fixed by the seed, photometric parameters are re-drawn at each level. Right: real Braille photographs ordered from clean
to degraded, for comparison.}
\label{fig:samples}
\end{figure*}

\begin{table*}[!tp]
\centering
\footnotesize
\setlength{\tabcolsep}{3pt}
\renewcommand{\arraystretch}{1.12}

\begin{tabularx}{\textwidth}{
    >{\raggedright\arraybackslash}p{2.3cm}
    >{\raggedright\arraybackslash}X
    >{\raggedright\arraybackslash}p{2.5cm}
    >{\centering\arraybackslash}p{1.2cm}
}
\toprule
Stage & Parameter and range & Prob. & Order \\
\midrule

Dot size &
dot radius = glyph radius $\times$ \texttt{dot\_scale}
$\times U(.88,1.12)\times U(.90,1.05)$;
\texttt{dot\_scale} $=$ target diameter / pitch / 0.68,
clipped to $[0.4,1.7]$ &
1.0 & 1 (render) \\

Dot jitter &
centre offset $\pm 2$\,px (scaled with font size);
dot intensity $U(200,255)$ &
1.0 & 1 \\

Height-map blur &
Gaussian radius $U(3,6)\times 0.3$\,px (scaled);
light direction $U(-1.5,1.5)^2\times U(.3,1)$;
relief gain $U(2.5,5)$ &
1.0 & 1 \\

Cell spacing &
cell pitch / dot pitch $R$:
60\% $\mathcal N(2.65,0.15)$,
else $U(2.15,3.3)$
($U(2.1,3.8)$ in add-on) &
1.0 & 1 \\

Word spacing &
blank advance $b$ (cell pitches):
85\% from measured real values
($-0.28$ calibration,
$\times$ log-normal $\sigma{=}0.1$;
floor $U(.15,.32)$),
else $U(.32,2.2)$;
add-on: $U(.3,2.5)$ &
1.0 & 1 \\

Line spacing &
line pitch $L$ (cell pitches):
real value $\times e^{\mathcal N(0,0.06)}$
in $[1.45,6]$;
add-on: $U(1.5,3.5)$ &
1.0 & 1 \\

Paper grain &
Gaussian grain $\sigma\sim U(4,12)$ &
0.70 & 2 \\

Illumination field &
smooth multiplicative field in $[0.75,1.25]$ &
1.0 & 3 \\

Dot shadows &
offset $\pm U\{1..4\}$\,px,
strength $U(0.2,0.5)$ &
1.0 & 4 \\

Stains &
1--4 blobs,
radii $U\{10..80\}\times U\{10..60\}$\,px,
$\alpha\sim U(.03,.15)$,
blur $\sigma\sim U(5,20)$ &
0.10 & 5 \\

Photo background &
multiply blend with one of 38 photos,
$\alpha =$ height map $\times U(.6,.95)$ &
0.30 & 6 \\

Perspective &
each corner moved by $\pm U(2,8)\%$
of the shorter side,
then cropped and resized &
0.40 & 7 \\

Page curl &
sinusoidal displacement,
amplitude $U(3,15)$\,px,
frequency $U(0.3,1.5)$,
axis $x$ or $y$ &
0.10 & 8 \\

Elastic warp &
random displacement
$\alpha\sim U(5,20)$\,px,
smoothing $\sigma\sim U(8,20)$ &
0.15 & 9 \\

In-plane rotation &
angle $U(-8^\circ,8^\circ)$
about the image centre
(canvas expanded, no clipping);
no separate shear is applied &
0.20 & 10 \\

Colour temperature &
warm / cool / neutral (1/3 each);
R or B gain $U(1.0,1.15)$,
opposite channel $U(0.85,1.0)$ &
0.60 & 11 \\

Low light &
brightness $\times U(0.35,0.70)$
plus noise $\sigma\sim U(8,20)$ &
0.15 (0.10)$^\dagger$ & 11 \\

Over-exposure &
edge band of $h/8$--$h/3$,
gain $U(1.2,1.8)$ $+U(20,80)$
(only if no low light) &
0.30 & 11 \\

Shadow bands &
1--2 linear bands,
strength $U(.2,.5)$,
width $U(.05,.25)$ &
0.15 & 11 \\

Noise block &
whole block applied with probability
0.40 (0.25)$^\dagger$ &
0.40 (0.25)$^\dagger$ & 12 \\

\quad down--up scaling &
scale factor $U(0.4,0.75)$
(bilinear down, bicubic up) &
0.5 & 12 \\

\quad salt \& pepper &
$U(0.02,0.15)\%$ of pixels each &
0.10 (0.02)$^\dagger$ & 12 \\

\quad Gaussian noise &
$\sigma\sim U(2,18)$ grey levels &
1.0 (in block) & 12 \\

\quad JPEG &
quality $U\{45..88\}$ &
1.0 (in block) & 12 \\

Coloured border &
4--36\,px border
(scaled to the cell pitch),
random colour &
0.05 & 13 \\

Anisotropic squeeze &
$x/y$ scale ratio
$e^{\mathcal N(0,0.10)}$,
clipped to $[1/1.45,1.45]$ &
1.0 & 14 (final) \\

Final blur / noise / JPEG &
at final resolution:
blur $\sigma\sim U(0,0.22)\times$ dot pitch;
noise $\sigma\sim U(.5,6)$;
JPEG $U\{30..90\}$ &
0.5 / 0.4$^\dagger$ / 0.5$^\dagger$ & 15 \\

\bottomrule
\end{tabularx}

\caption{
Augmentations: sampling ranges and per-image probability.
$U(a,b)$ is uniform, $\mathcal N$ Gaussian.
$\dagger$: add-on runs use the value in parentheses.
Camera-optics effects (defocus, motion blur, lens distortion,
chromatic aberration, vignetting), vertical flips, and
$90^\circ$ rotations are implemented but disabled ($p=0$).
}

\label{tab:aug}
\end{table*}

\begin{table*}[!tp]\centering\footnotesize

\begin{tabular}{p{3.6cm}p{7.3cm}p{3.5cm}}\toprule
Quantity & Sampling & Realised (p5--p95) \\\midrule
Cell pitch $P$ (px) & bank record $\times e^{\mathcal N(0,0.08)}$ (10\% of samples: $\sigma{=}0.45$), clipped to $[5,110]$; texts of $\ge$16 / $\ge$32 cells get a floor of 18 / 22\,px & 20--64 \\
Cell tightness $R$ & see Table~\ref{tab:aug} & 2.24--3.53 \\
Word gap $b$ & see Table~\ref{tab:aug} & 0.20--2.20 \\
Dot diameter / pitch & bank $\times 0.9\,e^{\mathcal N(0,0.08)}$ & 0.33--0.81 \\
$x/y$ squeeze & $e^{\mathcal N(0,0.10)}$ & 0.85--1.18 \\
Margins (cells) & bank $+\mathcal N(0,0.06)$; every side $\ge 0.08$ (re-drawn in $[0.08,0.23]$ otherwise); no dot is clipped (min.\ tracked dot-to-border distance 0.9\,px) & left 0.11--3.51 \\
Visual-token budget & $\le 1024$ Qwen visual tokens (2.8\% of images are scaled down) & 84--1008 tokens (p50--p99) \\\midrule
Embossed (default) & height-map shading, as above & 85.5\% \\
Tape strips & each line on a coloured strip (11 colours, gloss $\pm15\%$, drop shadow) over a page or photo; light embossed dots & 7.0\% \\
Ghost cells & flat print: six faint dots per cell (size $U(.22,.45)\times$ real dot, $\alpha\sim U(.6,1)$) under large dark dots; texts $\le4$ words & 4.3\% \\
Hollow dots & ring-shaped dots, inner radius $U(.45,.75)$ of the dot radius, 30\% with ghost cells & 3.2\% \\
Close-up & texts $\le10$ cells at $P\sim$ log-uniform $[60,140]$\,px; half with a blurry low-resolution upscale (factor $U(.2,.5)$) & 1.5\% \\
Dynamic spacing & $R$, $b$, $L$ uniform over wide ranges instead of the real bank & 31.4\% \\\bottomrule

\end{tabular}
\caption{Real-matched layout and the optional rendering styles. Distributions marked ``bank'' are resampled from measurements on real
braille photographs (cell pitch, dot size, margins, line count, line pitch, word gaps; $\approx$3k reliable records); the last column gives the realised
5th--95th percentile in the final corpus (256{,}717 real-matched images).}
\label{tab:layout}
\end{table*}

\section{Training Details}
\label{sec:training_hyperparameters}

Both Qwen3.5-2B and Chandra-OCR-2 are fully fine-tuned for one epoch on the
synthetic corpus with an image-to-text transcription objective, using
distributed training on 8 NVIDIA H100 GPUs.
Table~\ref{tab:training_hyperparameters} lists the hyperparameters. Where two values are given, the first is for
Qwen3.5-2B (4 gradient-accumulation steps, cosine schedule) and the second for Chandra-OCR-2 (8 steps, linear schedule).
For the primary full-fine-tuning results, the last checkpoint is used for evaluation. Continual-training analyses report the checkpoints at their stated numbers of samples seen, including the 15k checkpoint in the corpus comparison.

\begin{table}[!htbp]
    \centering
    \small
    \begin{tabular}{lc}
        \toprule
        \textbf{Hyperparameter} & \textbf{Value} \\
        \midrule
        Epochs & 1 \\
        Number of GPUs & 8 \\
        Batch size per GPU & 1 \\
        Gradient accumulation steps & {4,8} \\
        Effective batch size & {32,64} \\
        Learning rate & $2\times10^{-5}$ \\
        LoRA learning rate & $1\times10^{-4}$ \\

        LR scheduler & Cosine / Linear \\
        Warmup ratio & 0.03 \\
        Weight decay & 0.01 \\
        Max.\ gradient norm & 1.0 \\
        Precision & BF16 \\
        Random seed & 42 \\
        \bottomrule
    \end{tabular}
    \caption{Training hyperparameters used for fine-tuning Qwen3.5-2B and
    Chandra-OCR-2 on Braille transcription.}
    \label{tab:training_hyperparameters}
\end{table}

\section{Evaluation Details}
\label{app:evaluation}

\subsection{Prompting and Scoring}

All standalone models receive the same zero-shot prompt,
\texttt{Decode the Braille text shown in the image and output the
transcription.}, with no system prompt or in-context examples. When a model
returns an explanation alongside its answer, the final transcription is
extracted before scoring. For normalized match, both predictions and
references are normalized with Unicode NFKC, lowercased, and have runs of
whitespace collapsed to a single space and leading and trailing whitespace
removed. The same scoring
code is used for BrailleBench and for WMT validation.

For the 44 Claude Code Grade-2 pairs, recomputing CER and WER by pooling edit counts and reference lengths reproduces the reported values (0.2515 and 0.4583). Applying the stated normalization first gives 0.2490 and 0.4550. Word precision, recall, and F1 are reproduced by computing lowercase whitespace-token multiset overlap per image and taking the mean across images, with zero precision and F1 for an empty prediction. Edit similarity is reproduced as the mean of $1-d/\max(|r|,|p|)$ over normalized reference/prediction pairs, where $d$ is their Levenshtein distance. ChrF is reproduced as the mean of raw-string sentence scores from \texttt{SacreBLEU 2.6.0}, using character order 6, word order 0, $\beta=2$, mixed case, and whitespace excluded; this gives 79.11. This reproduction establishes these settings for the audited Claude Code Grade-2 row; scoring configurations for other rows require their evaluation code.

\subsection{Tool-Assisted Harnesses}
\label{app:harness}

We run GPT-5.6 Sol through the Codex harness and Claude Opus 5.5 (high
effort) through Claude Code. In both settings the agent can run shell commands,
including image-processing tools, and iterate over several views of the image before
giving its final transcription; in practice only Codex also wrote and ran its own
code (Table~\ref{tab:harness-tools}). Each image takes approximately 180 seconds on
average. Table~\ref{tab:harness-cost} reports token usage and
API-equivalent cost over all 259 BrailleBench images. Claude Code uses
about 1.42M tokens per image (\$0.58 per image), and Codex about 0.70M
tokens per image (\$0.51 per image).

\paragraph{How the harnesses decode an image.}
We reconstructed each harness's behaviour from its tool-call log by assigning
every call to the image being processed (Table~\ref{tab:harness-tools}).
In Claude Code, no tool ever produced a Braille character. The tools only
produced enlarged and cleaned-up views of the image; the model read the dots
from these views and wrote the Unicode transcription itself. Each image
followed the same loop. The agent first checked the image size and looked at
the original or a 2--3$\times$ upscaled copy. It then cropped the image to a
single line or to segments of about 5--10 cells, enlarged them 4--12$\times$,
and applied filters only when the dots were not clearly visible. It read each
view, re-cropped or changed the filter when dots were clipped or unclear,
decoded the cells, and appended the prediction. The most frequent operations
were upscaling (92\% of images), grayscale conversion (58\%), cropping to a
line or segment (55\%) and contrast normalization (52\%). High-pass filtering
for embossed paper (12\%), rotation or deskewing (10\%) and thresholding (4\%)
were used on harder images; dot-coordinate measurement was used on 3\%.
Effort was highly uneven: 119 images needed a single processed view, while
multi-line pages took up to 62 tool calls. The decoding decision itself
happens in the model's reasoning, which is not stored in readable form, so the
logs show which views the model looked at but not how it read each cell.
Codex makes a similar number of tool calls per image (Table~\ref{tab:harness-tools}).

\begin{table}[!htbp]
\centering
\small
\setlength{\tabcolsep}{3pt}
\renewcommand{\arraystretch}{1.12}
\begin{tabular}{@{}lrrrc@{}}
\toprule
Harness & \shortstack{Tool\\calls} &
\shortstack{Image\\views} &
\shortstack{Image\\commands} &
\shortstack{Code\\run} \\
\midrule
Claude Code & 6.8 & 3.2 & 3.5 & No \\
Codex       & 7.2 & 4.1 & 3.7 & Yes \\
\bottomrule
\end{tabular}
\caption{Mean tool use per image in the tool-assisted
harnesses. Image views are processed crops read by
the model; image commands are image-processing
invocations (ImageMagick for Claude Code).}
\label{tab:harness-tools}
\end{table}

\begin{table}[!htbp]
\centering
\small
\resizebox{\columnwidth}{!}{%
\begin{tabular}{@{}lrrrr@{}}
\toprule
Harness & Images & Tokens &
\shortstack{Cost\\(USD)} &
\shortstack{USD /\\image} \\
\midrule
Claude Code & 259 & 367{,}395{,}129 & 150.82 & 0.58 \\
Codex       & 259 & 180{,}338{,}976 & 132.87 & 0.51 \\
\bottomrule
\end{tabular}%
}
\caption{Token usage and API-equivalent cost of the tool-assisted
harnesses on BrailleBench.}
\label{tab:harness-cost}
\end{table}

\section{Error Analysis Counts}
\label{app:error_counts}

Table~\ref{tab:error-counts} gives the exact counts behind Figure~\ref{fig:error_breakdown}. Categories are mutually
exclusive, so rows sum to the number of images in the grade (215 for Grade-1, 44 for Grade-2). Exact-match counts
correspond to the exact-match rates in Table~\ref{tab:braille-all-models}. \emph{No output} covers empty outputs and
refusals, and also two Gemma 4 images for which no response was returned. These are counts of textual outcomes; they do not
identify the visual cause of an error.

\paragraph{Recomputed Claude Code Grade-2 categories.}
For this row, we apply the following deterministic rules in order: raw string equality gives exact match; an empty normalized prediction gives no output; equality after Unicode NFKC, lowercasing, and collapsing and trimming whitespace gives formatting only; equality after additionally removing all whitespace gives spacing only. Remaining predictions are partial errors when their character-level Levenshtein distance is at most half the normalized reference length, and large errors otherwise. Punctuation is retained. The supplied pairs contain one empty prediction and no explicit refusal. This protocol yields 18 exact, 0 formatting-only, 9 spacing-only, 8 partial, 8 large, and 1 no-output case. The original looser formatting transform used for the other rows is not fully specified, so comparisons of these error-type splits with the recomputed row are provisional. Exact-match comparisons are unaffected by that normalization difference.

\begin{table*}[!tp]
\centering
\small
\setlength{\tabcolsep}{4pt}
\renewcommand{\arraystretch}{1.1}
\begin{tabular}{@{}l rrrrrr @{\hspace{14pt}} rrrrrr@{}}
\toprule
& \multicolumn{6}{c}{\textbf{Grade-1} ($n=215$)} & \multicolumn{6}{c}{\textbf{Grade-2} ($n=44$)} \\
\cmidrule(lr){2-7}\cmidrule(l){8-13}
Model & E & F & S & P & L & N & E & F & S & P & L & N \\
\midrule
Claude Code (Opus 5.5) & 190 & 14 & 1 & 9 & 1 & 0 & 18 & 0 & 9 & 8 & 8 & 1 \\
Qwen3.5-2B (ours) & 187 & 6 & 4 & 16 & 2 & 0 & 27 & 5 & 1 & 10 & 1 & 0 \\
Chandra-OCR-2 (ours) & 183 & 7 & 1 & 20 & 4 & 0 & 24 & 6 & 0 & 13 & 1 & 0 \\
Codex (GPT-5.6 Sol) & 135 & 20 & 0 & 22 & 6 & 32 & 0 & 0 & 0 & 1 & 42 & 1 \\
GPT-6.1 Sol & 116 & 8 & 2 & 35 & 24 & 30 & 5 & 1 & 0 & 7 & 9 & 22 \\
GPT-5.6 Sol High & 24 & 10 & 0 & 13 & 166 & 2 & 2 & 1 & 0 & 7 & 31 & 3 \\
Angelina & 7 & 3 & 1 & 88 & 31 & 85 & 0 & 0 & 0 & 25 & 14 & 5 \\
Gemini 3.5 Flash & 6 & 5 & 0 & 22 & 182 & 0 & 0 & 2 & 0 & 6 & 34 & 2 \\
Claude Opus 4.8 & 2 & 1 & 0 & 12 & 194 & 6 & 0 & 1 & 0 & 4 & 33 & 6 \\
Gemma 4 31B-IT & 1 & 1 & 0 & 7 & 205 & 1 & 0 & 0 & 1 & 2 & 36 & 5 \\
GPT-5.5 & 0 & 6 & 0 & 11 & 155 & 43 & 0 & 0 & 0 & 5 & 39 & 0 \\
Kimi K2.6 & 0 & 0 & 0 & 1 & 211 & 3 & 0 & 0 & 0 & 0 & 42 & 2 \\
\bottomrule
\end{tabular}
\caption{Error-analysis counts per system on BrailleBench. E: exact match; F: formatting only; S: spacing only;
P: partial error; L: large error; N: no output. Claude Code Grade-2 is recomputed from 44 prediction/reference pairs using NFKC, lowercase, and whitespace normalization with punctuation retained; its counts sum to 44. Other rows retain the supplied categorization under the original, looser formatting normalization.}
\label{tab:error-counts}
\end{table*}

\section{Qualitative Examples of Harness Failures}
\label{app:harness_failures}

Although the harnesses are the strongest systems on Grade-1, their errors
are often fluent rather than visibly broken.
Figure~\ref{fig:combined-qualitative} shows four examples. Claude Code
tends to replace the text with a plausible alternative, such as a similar
word or a sentence from elsewhere in the same story, while Codex more often
produces strings of unrelated characters. Fluent errors of this kind are
hard to detect for a user who cannot read Braille.

\tcbset{
  hbpanel/.style={enhanced, colback=white, colframe=black!35,
    colbacktitle=black!6, coltitle=black, fonttitle=\bfseries\small,
    boxrule=0.5pt, arc=2pt, left=4pt, right=4pt, top=4pt, bottom=4pt,
    toptitle=2pt, bottomtitle=2pt, before skip=0pt, after skip=0pt},
  hbinner/.style={enhanced, frame hidden, arc=1.5pt,
    left=4pt, right=4pt, top=3pt, bottom=3pt, before skip=0pt, after skip=0pt},
}

\newcommand{\hbInput}[3]{%
  \begin{tcolorbox}[hbpanel, title={#1\enspace Input}, equal height group=hbInput]
    \begin{tcolorbox}[hbinner, colback=green!6]
      \small\textbf{Prompt:} \texttt{<image>} Decode the Braille text shown
      in the image and output the transcription.\par\smallskip
      \parbox[c][1.5cm][c]{\linewidth}{\centering
        \includegraphics[width=0.9\linewidth,height=1.5cm,keepaspectratio]{#2}}
    \end{tcolorbox}\par\smallskip
    \begin{tcolorbox}[hbinner, colback=black!4, equal height group=hbGT]
      \small\textbf{Ground truth:} \texttt{#3}
    \end{tcolorbox}
  \end{tcolorbox}}

\newcommand{\hbFail}[2]{%
  \begin{tcolorbox}[hbpanel, equal height group=hbResult,
      title={#1 -- \textcolor{red!70!black}{Failure}}]
    \begin{tcolorbox}[hbinner, colback=red!7, equal height group=hbResp]
      \small\textbf{Response:}\par
      {\ttfamily\raggedright #2\par}
      \hfill\textcolor{red!70!black}{\ding{55}}
    \end{tcolorbox}
  \end{tcolorbox}}

\newcommand{\hbOk}[2]{%
  \begin{tcolorbox}[hbpanel, equal height group=hbResult,
      title={#1 -- \textcolor{green!45!black}{Success}}]
    \begin{tcolorbox}[hbinner, colback=green!8, equal height group=hbResp]
      \small\textbf{Response:}\par
      {\ttfamily\raggedright #2\par}
      \hfill\textcolor{green!45!black}{\ding{51}}
    \end{tcolorbox}
  \end{tcolorbox}}

\newcommand{\hbExample}[5]{%
  \begin{minipage}[t]{0.48\textwidth}
    \hbInput{#1}{#2}{#3}\par\vspace{4pt}
    \begin{minipage}[t]{0.49\linewidth}#4\end{minipage}\hfill
    \begin{minipage}[t]{0.49\linewidth}#5\end{minipage}
  \end{minipage}}

\begin{figure*}[!tp]
  \centering

  \hbExample{(a)}{}{mother's Room}
    {\hbFail{Claude Code}{men's room}}
    {\hbFail{Codex}{ich''-'' / shm y-xxowstk}}%
  \hfill
  \hbExample{(b)}{}{Now Lucy knows the secret code.}
    {\hbFail{Claude Code}{\textquotedbl Unlucky,Lucy says, \textquotedbl I don't}}
    {\hbFail{Codex}{;inowtheustyl / st-l'';amthb}}

  \vspace{10pt}

  \hbExample{(c)}{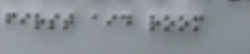}{first aid room}
    {\hbFail{Claude Code}{prison ain heem}}
    {\hbOk{Codex}{first aid room}}%
  \hfill
  \hbExample{(d)}{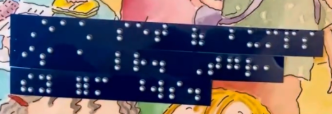}{Each set of bumps is a letter. Together, they form words.}
    {\hbFail{Claude Code}{``Now you try,'' he says.Lucy closes her eyesand feels the dots.}}
    {\hbFail{Codex}{yv''ypxacshk / zqjhpkewl-k}}

  \caption{Examples from the tool-assisted harnesses on BrailleBench.
  (a)~Claude Code (Opus 5.5) misreads ``mother's'' as ``men's'', while Codex
  generates unrelated characters. (b)~Claude Code produces story text that is
  not in the image, while Codex again produces unrelated characters.
  (c)~On a blurred, low-resolution sign, Claude Code outputs a fluent but wrong
  phrase, while Codex transcribes it correctly. (d)~Claude Code outputs a
  sentence from a different page of the same book, while Codex produces
  unrelated characters.}
  \label{fig:combined-qualitative}
\end{figure*}

\section{Synthetic-to-Real Analyses}
\label{app:syn_real}

This section extends Section~\ref{subsec:syn_to_real}. We first study how corpus size and text length affect transfer,
then give the full trajectory of the continual-training experiment, and finally report the full LoRA-rank results.

\subsection{Corpus Size and Text Length}
\label{app:wmt_eval}

Our main model is trained on the 276k corpus. We compare it with two variants: Qwen3.5-2B trained on the 500k corpus,
which is nearly twice as large and has about four times longer texts (Table~\ref{tab:training_corpus_profiles}), and the
500k model after continual LoRA training on 15k samples from the 276k corpus. We evaluate all three on BrailleBench and on
WMT validation (Section~\ref{subsec:syn_real_setup}), which none of them sees during training. WMT validation sentences
are much longer than BrailleBench texts (81.5 vs.\ 22.2 characters on average), so the two sets measure reading short
real-world text and long synthetic text, respectively. Table~\ref{tab:corpus_variants} reports the results.

\begin{table}[!htbp]
\centering
\scriptsize
\setlength{\tabcolsep}{3pt}
\renewcommand{\arraystretch}{1.1}
\begin{tabular}{lrrr}
\toprule
Metric & 500k & 276k & WMT-Braille \\
\midrule
Records & 500,000 & 276,717 & 100,000 \\
\shortstack[l]{Characters\\mean/med./max}
 & \shortstack{148.1/150\\300}
 & \shortstack{39.1/21\\160}
 & \shortstack{82.9/76\\200} \\
\shortstack[l]{Words\\mean/med./max}
 & \shortstack{25.4/25\\50}
 & \shortstack{6.9/4\\49}
 & \shortstack{13.8/13\\43} \\
\shortstack[l]{Width (px)\\mean/med./range}
 & \shortstack{1536/1857\\172--2530}
 & \shortstack{631/493\\28--2301}
 & \shortstack{604/600\\488--715} \\
\shortstack[l]{Height (px)\\mean/med./range}
 & \shortstack{706/623\\11--2103}
 & \shortstack{279/137\\28--2048}
 & \shortstack{242/225\\90--708} \\
\bottomrule
\end{tabular}
\caption{Text and image size statistics of our two training corpora (500k, 276k) and the prior work's WMT-Braille data (WMT-Train and WMT validation together). Statistics cover every record.}
\label{tab:training_corpus_profiles}
\end{table}

\begin{table}[!htbp]
\centering
\small
\setlength{\tabcolsep}{3pt}
\renewcommand{\arraystretch}{1.1}
\resizebox{\columnwidth}{!}{%
\begin{tabular}{l rrr @{\hspace{8pt}} rrr}
\toprule
& \multicolumn{3}{c}{BrailleBench ($n=259$)} & \multicolumn{3}{c}{WMT val.\ ($n=1{,}000$)} \\
\cmidrule(lr){2-4}\cmidrule(l){5-7}
Training data & chrF & EM (\%) & CER & chrF & EM (\%) & CER \\
\midrule
276k (short texts)        & \textbf{94.97} & \textbf{82.63} & \textbf{0.0305} & 87.14 & 41.70 & 0.6331 \\
500k (long texts)         & 87.97 & 72.97 & 0.1981 & \textbf{95.25} & 63.10 & \textbf{0.0345} \\
500k $+$ 15k short (LoRA) & 92.41 & 81.85 & 0.0640 & 94.15 & \textbf{70.20} & 0.0399 \\
\bottomrule
\end{tabular}%
}
\caption{Qwen3.5-2B trained on our two synthetic corpora, and the 500k model after continual LoRA training on 15k samples
from the 276k corpus. BrailleBench covers both grades; WMT validation is synthetic and not seen in training. Higher chrF and
exact match (EM) and lower CER are better; the best value per column is bold.}
\label{tab:corpus_variants}
\end{table}

\paragraph{Does more, longer synthetic data help?}
No. Scaling up to the 500k corpus lowers BrailleBench exact match from 82.63\% to 72.97\% and raises CER from 0.0305 to
0.1981. On WMT validation the pattern reverses: exact match rises from 41.70\% to 63.10\% and CER falls from 0.6331 to
0.0345. Each model is best on the evaluation set whose text length is closer to its training data, which is consistent with an effect of text length. The 500k corpus also has larger images and more samples, so these factors cannot be fully
separated. The 276k model's high WMT CER despite a chrF of 87.14 suggests that a minority of its outputs contain long
insertions, since CER is unbounded but chrF is not. This comparison also shows how misleading synthetic validation can be:
selected on WMT validation alone, the 500k model would be chosen, although it is clearly worse on real Braille.

\paragraph{Can short samples repair the 500k model?}
If the 500k model is weaker on BrailleBench because it has seen too few short texts, a small amount of short-text training
should recover its accuracy. After continual LoRA training on 15k samples from the 276k corpus, BrailleBench exact match
rises from 72.97\% to 81.85\% and CER falls from 0.1981 to 0.0640, which supports the text-length explanation. The model
does not lose its long-text ability: WMT exact match also rises, to 70.20\%, while chrF drops only slightly. It nearly matches
the 276k model on BrailleBench and is far better on long texts (70.20\% vs.\ 41.70\%). The 276k model was already our main model before these analyses were run, so the 500k variants are reported
as analyses only; their BrailleBench CER also remains about twice that of the 276k model.

\subsection{Continual Training Trajectory}
\label{app:continual_training}

\begin{figure*}[!tp]
    \centering
    \includegraphics[width=\textwidth]
        {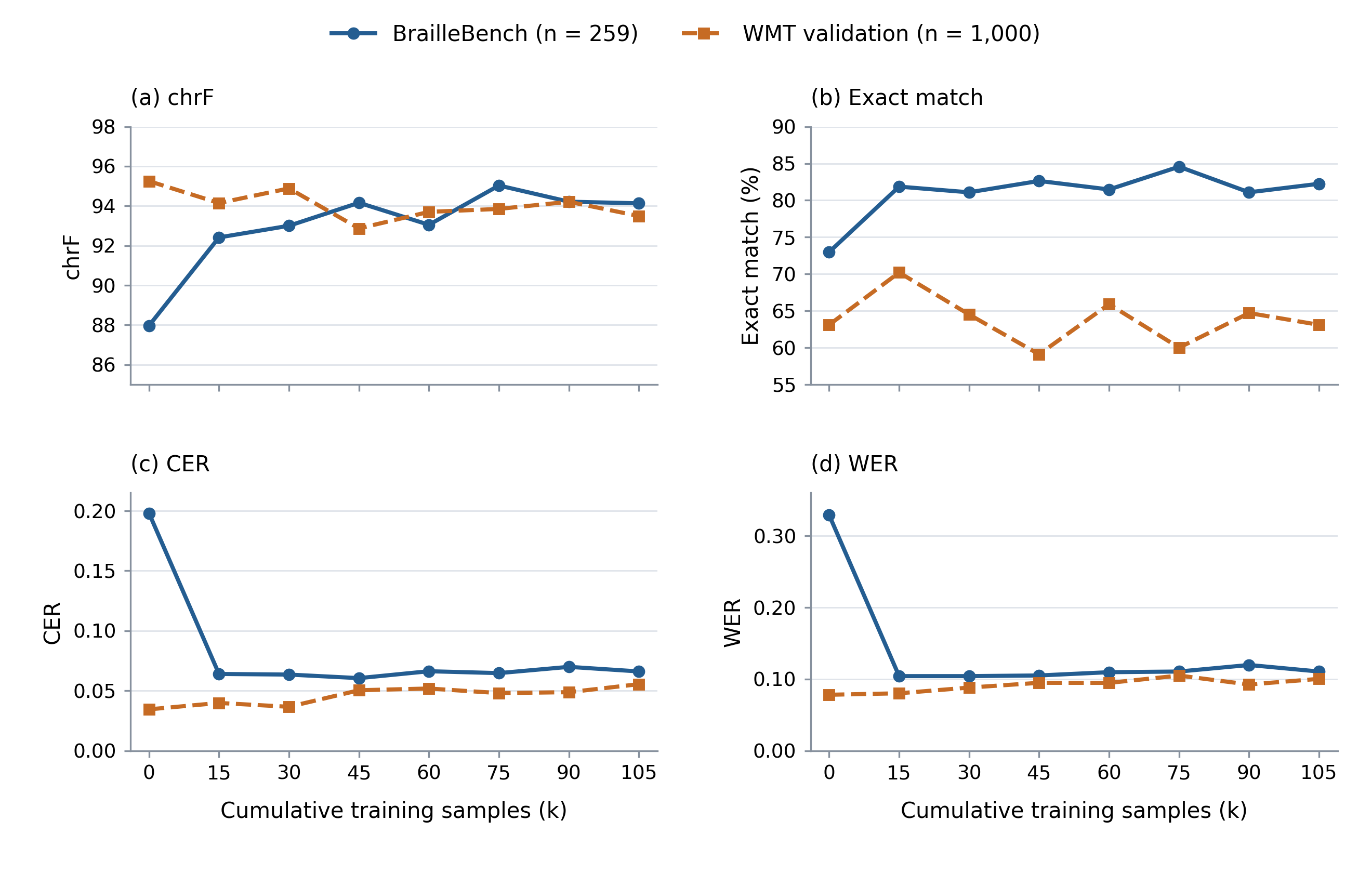}
    \caption{Qwen3.5-2B performance during continual LoRA
    training on samples from our 276k corpus, following
    pretraining on 500k longer synthetic samples.}
    \label{fig:continual_training_scaling}
\end{figure*}

Figure~\ref{fig:continual_training_scaling} extends the 0k--15k comparison above to checkpoints after 30k, 45k, 60k, 75k,
90k, and 105k cumulative samples from the same single run. Every checkpoint is evaluated on all 259 BrailleBench images and
the 1,000 WMT validation images.

\paragraph{BrailleBench.}
Almost all of the improvement happens within the first 15k samples. Later gains are small and uneven: the 75k checkpoint
has the highest exact match (84.56\%) and chrF (95.03), CER is lowest at 45k (0.0606), and WER is lowest at 15k and 30k
(0.1042). At the final 105k checkpoint, exact match is 82.24\% and CER 0.0661.

\paragraph{WMT validation.}
WMT exact match peaks at 15k (70.20\%) and returns to its starting value of 63.10\% at 105k, while chrF, CER, and WER are
best before continual training. Gains on real Braille therefore do not coincide with gains on synthetic validation, even
though both measure the same task. As in Section~\ref{subsec:syn_to_real}, synthetic validation scores are not a reliable
proxy for real-image accuracy, so checkpoints should not be selected on them alone.

\subsection{Effect of LoRA Rank}
\label{app:lora_rank}

\begin{figure*}[!tp]
    \centering
    \includegraphics[width=\textwidth]
        {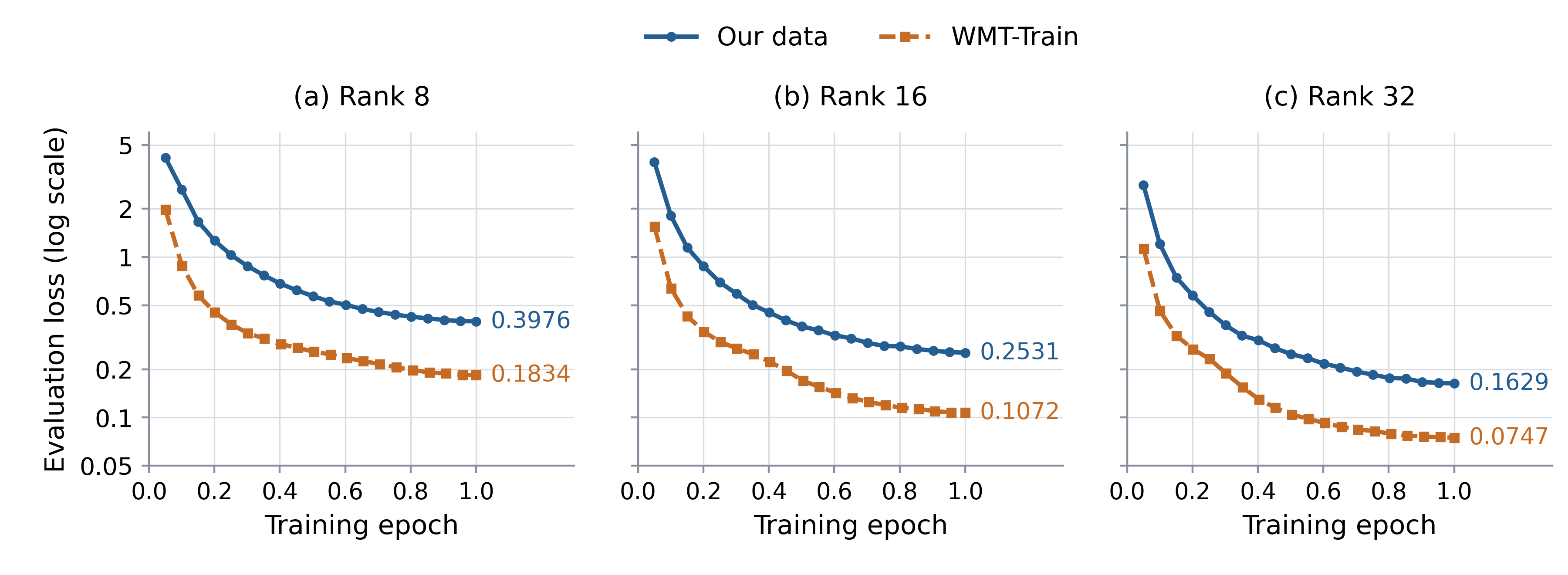}
    \caption{Evaluation loss of Llama-3.2-11B-Vision-Instruct
    during one epoch of LoRA training at ranks 8, 16, and 32,
    on our 276k corpus (solid) and on WMT-Train (dashed).
    Lower WMT-Train loss does not mean better transcription:
    the two datasets differ in text length and difficulty.}
    \label{fig:lora_rank_loss}
\end{figure*}

Section~\ref{subsec:syn_to_real} shows that Llama-3.2-11B-Vision-Instruct trained with LoRA on our data reaches 72--76\%
exact match on BrailleBench at every rank. Here we give the training curves and the WMT validation scores of these runs.
We train at ranks 8, 16, and 32 for one epoch, either on our 276k corpus (4,309 steps) or on WMT-Train (1,547 steps).
Higher ranks reach lower evaluation loss throughout training on both datasets (Figure~\ref{fig:lora_rank_loss}): on our
data, the final loss falls from 0.3976 at rank 8 to 0.2531 at rank 16 and 0.1629 at rank 32, and on WMT-Train from 0.1834 to
0.1072 and 0.0747.

\begin{table}[!htbp]
\centering
\small
\setlength{\tabcolsep}{3pt}
\renewcommand{\arraystretch}{1.1}
\begin{tabular}{rrrrr}
\toprule
Rank & chrF $\uparrow$ & EM (\%) $\uparrow$
     & CER $\downarrow$ & WER $\downarrow$ \\
\midrule
\multicolumn{5}{l}{\textit{BrailleBench ($n=259$)}} \\
8  & 88.28 & 72.59 & 0.4196 & 1.0704 \\
16 & 88.57 & 72.20 & \textbf{0.0754} & 0.1444 \\
32 & \textbf{90.10} & \textbf{76.06}
   & 0.0759 & \textbf{0.1261} \\
\midrule
\multicolumn{5}{l}{\textit{WMT validation ($n=1{,}000$)}} \\
8  & 95.27 & 73.10 & 0.0595 & 0.0450 \\
16 & 92.43 & 53.90 & 0.0364 & 0.0783 \\
32 & \textbf{96.63} & \textbf{81.60}
   & \textbf{0.0161} & \textbf{0.0306} \\
\bottomrule
\end{tabular}
\caption{Transcription scores of Llama-3.2-11B-Vision-Instruct
trained with LoRA on our 276k corpus, across LoRA ranks. Best scores within each evaluation set
are bold.}
\label{tab:lora_rank_results}
\end{table}

Lower loss does not always mean better transcription (Table~\ref{tab:lora_rank_results}). On BrailleBench, rank 32 has the
highest exact match (76.06\%) and chrF and the lowest WER, while CER is almost the same at ranks 16 and 32. Rank 8 produces
2,057 words against 1,094 reference words, which explains its high CER and WER despite a similar exact match. On WMT
validation, rank 32 is best on all four metrics (81.60\% exact match), but rank 16 is worse than rank 8 in exact match, chrF,
and WER despite its lower loss. Across all ranks, error rates remain higher on BrailleBench than on WMT validation,
reflecting the greater difficulty of reading real images.

\section{Out-of-Benchmark Evaluation}
\label{app:qualitative}

\subsection{Children's Braille Book}

We additionally evaluate on 142 images from a single Grade-1 Braille
children's book. These images are \emph{not} part of BrailleBench: because
they all come from one source, including them would let a single book
dominate the benchmark. Table~\ref{tab:braille-book-ocr} reports aggregate
scores, and Table~\ref{tab:qualitative-braille-comparison} shows three
examples. Our models clearly outperform the Angelina Braille Reader
\cite{ovodov2021optical}, which frequently introduces substitutions,
omissions, and uncertain characters (e.g., 94.37\% exact match for
Chandra-OCR-2 versus 0\% for Angelina). The Llama model trained on the prior
work's WMT-Train data produces no correct transcriptions, which is consistent
with its BrailleBench results. As on BrailleBench, the 276k Qwen model
outperforms the 500k Qwen model, whose training texts are much longer.

\begin{table*}[!tp]
\centering
\small
\setlength{\tabcolsep}{8pt}
\renewcommand{\arraystretch}{1.18}

\begin{tabular*}{\textwidth}{
@{\extracolsep{\fill}}lccccc@{}
}
\toprule

\textbf{Metric} &
\textbf{Angelina OBR} &
\shortstack{\textbf{Chandra-OCR-2}\\\textbf{(ours)}} &
\shortstack{\textbf{Qwen3.5-2B}\\\textbf{500k (ours)}} &
\shortstack{\textbf{Qwen3.5-2B}\\\textbf{276k (ours)}} &
\shortstack{\textbf{Llama-3.2-11B}\\\textbf{(WMT-Train)}} \\

\midrule

CER $\downarrow$
& 0.7030 & \textbf{0.0138} & 0.1046 & 0.0218 & 6.7342 \\

Character accuracy $\uparrow$
& 0.2970 & \textbf{0.9862} & 0.8954 & 0.9782 & 0.0000 \\

WER $\downarrow$
& 1.0159 & \textbf{0.0438} & 0.1713 & 0.0518 & 15.6175 \\

Word accuracy $\uparrow$
& 0.0000 & \textbf{0.9562} & 0.8287 & 0.9482 & 0.0000 \\

\midrule

Normalized CER $\downarrow$
& 0.5650 & \textbf{0.0109} & 0.1017 & 0.0211 & 6.6754 \\

Normalized WER $\downarrow$
& 0.8725 & \textbf{0.0359} & 0.1633 & 0.0478 & 15.6175 \\

Exact match $\uparrow$
& 0.0000 & \textbf{0.9437} & 0.8028 & 0.9366 & 0.0000 \\

Normalized exact match $\uparrow$
& 0.0000 & \textbf{0.9577} & 0.8099 & 0.9437 & 0.0000 \\

ChrF $\uparrow$
& 13.99 & 98.43 & 91.21 & \textbf{98.98} & 2.70 \\

Edit similarity $\uparrow$
& 0.2872 & \textbf{0.9917} & 0.9380 & 0.9912 & 0.0875 \\

\midrule

Word precision $\uparrow$
& 0.1377 & 0.9730 & 0.8640 & \textbf{0.9777} & 0.0008 \\

Word recall $\uparrow$
& 0.1367 & 0.9718 & 0.8628 & \textbf{0.9730} & 0.0016 \\

Word F1 $\uparrow$
& 0.1370 & 0.9723 & 0.8622 & \textbf{0.9747} & 0.0010 \\

\bottomrule

\end{tabular*}

\caption{
Transcription scores on 142 images from a Grade-1
children's book that is not part of BrailleBench.
Bold indicates the best value for each metric.
$\downarrow$: lower is better;
$\uparrow$: higher is better.
CER: character error rate;
WER: word error rate.
ChrF is reported on a 0--100 scale;
all other scores retain their reported decimal values.
}

\label{tab:braille-book-ocr}

\end{table*}

\begin{table*}[!tp]
\centering

\setlength{\tabcolsep}{5pt}
\renewcommand{\arraystretch}{1.22}

\begin{tabularx}{\textwidth}{
    >{\centering\arraybackslash}m{0.245\textwidth}
    >{\raggedright\arraybackslash}m{0.205\textwidth}
    >{\raggedright\arraybackslash}X
    >{\raggedright\arraybackslash}X
}
\toprule

\rowcolor{headerbg}
\textbf{Braille Input}
&
\textbf{Ground Truth}
&
\textbf{Angelina Braille Reader}
&
\textbf{Qwen3.5-2B (Ours)}
\\

\midrule

\includegraphics[
    width=\linewidth,
    height=1.55cm,
    keepaspectratio
]{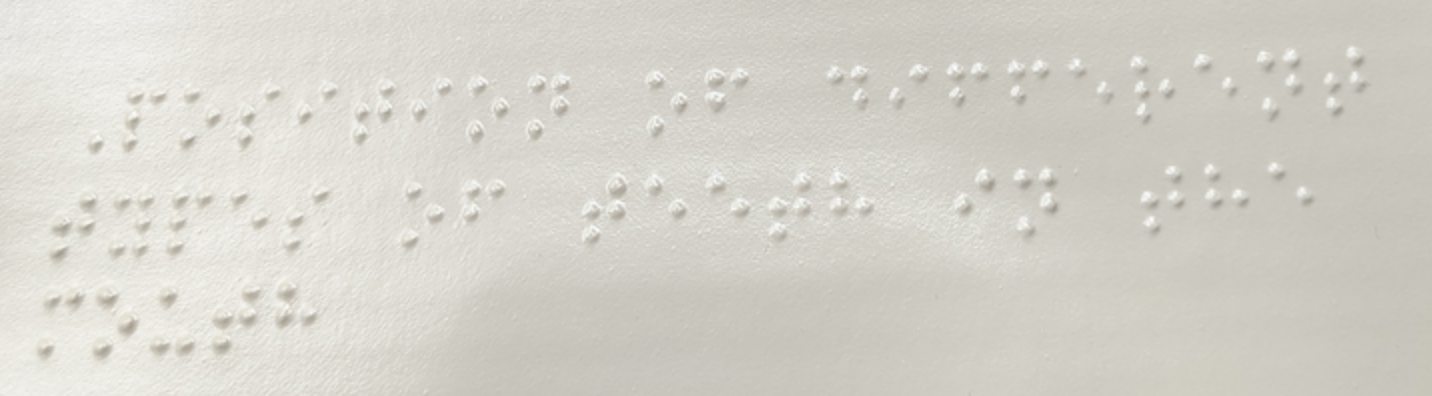}
&
Position of different types of teeth in the mouth
&
\cellcolor{failurebg}
\xmark\;
\texttt{\detokenize{sosition of di(fereat}}\par
\texttt{\detokenize{types of teeth in the}}\par
\texttt{\detokenize{m~~?~ut~?~~}}
&
\cellcolor{successbg}
\cmark\;
\textbf{Position of different types of teeth in the mouth}
\\

\midrule

\includegraphics[
    width=\linewidth,
    height=1.25cm,
    keepaspectratio
]{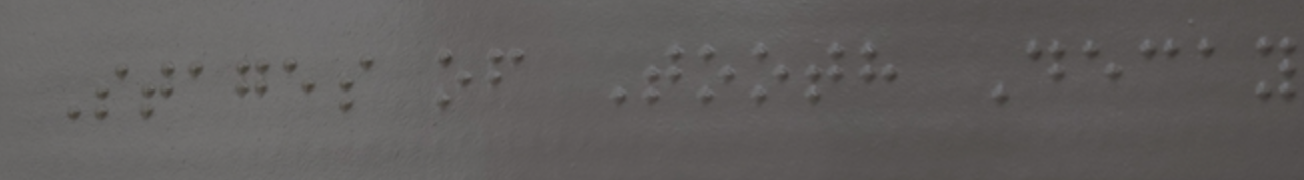}
&
Stages of Tooth Decay
&
\cellcolor{failurebg}
\xmark\;
\texttt{\detokenize{sfages of t»oth dec~?~y}}
&
\cellcolor{successbg}
\cmark\;
\textbf{Stages of Tooth Decay}
\\

\midrule

\includegraphics[
    width=\linewidth,
    height=1.25cm,
    keepaspectratio
]{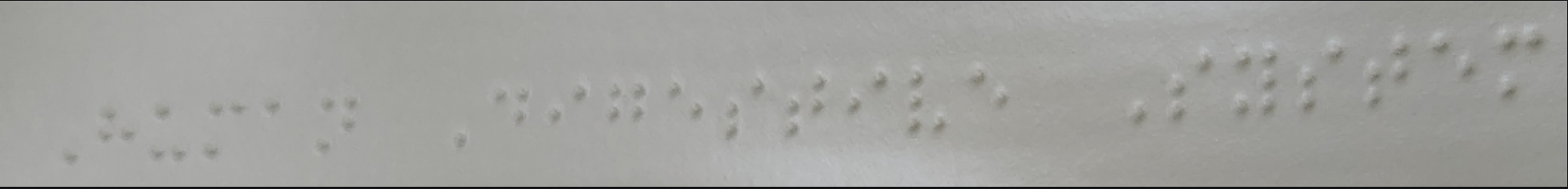}
&
Human Digestive System
&
\cellcolor{failurebg}
\xmark\;
\texttt{\detokenize{h~?~man digestiv syste?}}
&
\cellcolor{successbg}
\cmark\;
\textbf{Human Digestive System}
\\

\bottomrule
\end{tabularx}

\vspace{2pt}

{\footnotesize
\textit{Note:} \cmark\ denotes an exact match with the reference transcription,
while \xmark\ indicates a recognition error. Uncertainty markers in the
Angelina outputs are retained as produced by the reader.
}
\caption{
\textbf{Qualitative comparison of Braille recognition on a Grade-1 Braille children's book.}
The Angelina Braille Reader frequently introduces substitutions,
omissions, and uncertain characters, whereas our fine-tuned
Qwen3.5-2B model exactly recovers the reference text across these examples.
}
\label{tab:qualitative-braille-comparison}
\end{table*}

\end{document}